\documentclass{article}
\PassOptionsToPackage{table}{xcolor}
\usepackage{iclr2027_conference,times}
\usepackage[lining,nosfdefault]{raleway}

\usepackage{graphicx}
\usepackage{wrapfig}
\usepackage{needspace}
\usepackage{amsmath,amssymb}
\usepackage{booktabs}
\usepackage{multirow}
\usepackage{caption}
\usepackage{xcolor}
\usepackage[pagebackref,breaklinks]{hyperref}
\usepackage{url}
\usepackage{microtype}
\AtBeginDocument{\microtypesetup{activate=false}}

\ExplSyntaxOn
\seq_new:N \l_paper_bib_words_seq
\tl_new:N \l_paper_bib_tail_tl
\cs_new_protected:Npn \paper_bib_keep_tail:n #1
  {
    \seq_set_split:Nnn \l_paper_bib_words_seq { ~ } {#1}
    \tl_clear:N \l_paper_bib_tail_tl
    \hbox_set:Nn \l_tmpa_box {}
    \bool_while_do:nn
      { ! \seq_if_empty_p:N \l_paper_bib_words_seq &&
        \dim_compare_p:nNn {\box_wd:N \l_tmpa_box} < {0.20\linewidth} }
      {
        \seq_pop_right:NN \l_paper_bib_words_seq \l_tmpa_tl
        \tl_put_left:Nx \l_paper_bib_tail_tl
          { \exp_not:V \l_tmpa_tl \c_space_tl }
        \hbox_set:Nn \l_tmpa_box {\tl_use:N \l_paper_bib_tail_tl}
      }
    \tl_trim_spaces:N \l_paper_bib_tail_tl
    \seq_use:Nn \l_paper_bib_words_seq { ~ } ~
    \dim_compare:nNnTF {\box_wd:N \l_tmpa_box} > {0.45\linewidth}
      {\tl_use:N \l_paper_bib_tail_tl}
      {\mbox{\tl_use:N \l_paper_bib_tail_tl}}
  }
\cs_new_eq:NN \KeepBibliographyTail \paper_bib_keep_tail:n
\ExplSyntaxOff

\definecolor{papertealtint}{HTML}{EDF5F3}
\definecolor{paperbluetint}{HTML}{EAF1F8}
\newcommand{\poseerrorheader}{Pos. (cm) / Ori. ($^\circ$)}
\newcommand{\papertablestyle}{%
  \small
  \setlength{\tabcolsep}{3pt}%
  \renewcommand{\arraystretch}{1.14}}

\newcommand{\efficiencyfigure}{figures/efficiency.pdf}

\iclrfinalcopy
  \newcommand{\demowebsite}{opendrivelab.com/EgoHumanoid-V2}

\definecolor{qwenpurple}{HTML}{5843C9}
\definecolor{citationblue}{rgb}{0.21,0.49,0.74}
\hypersetup{
  colorlinks=true,
  linkcolor=red,
  citecolor=citationblue,
  urlcolor=citationblue
}

\title{\raggedright EgoHumanoid-V2: Human-to-Humanoid \\
Transfer of Coordinated Whole-Body \\
Skills for Loco-Manipulation}

\author{%
    \parbox[t]{\dimexpr\textwidth-2\tabcolsep\relax}{\raggedright
      {\bfseries
      \mbox{Jin Chen\textsuperscript{1,2,3,4,\textdagger}}\quad
      \mbox{Yiming Jiang\textsuperscript{2,5}}\quad
      \mbox{Chongyang Xu\textsuperscript{2,6}}\quad
      \mbox{Modi Shi\textsuperscript{7}}\quad
      \mbox{Shijia Peng\textsuperscript{7}}\quad
      \mbox{Li Chen\textsuperscript{7}}\quad
      \mbox{Tianyu Li\textsuperscript{7}}\quad
      \mbox{Mu Xu\textsuperscript{2}}\quad
      \mbox{Yilun Chen\textsuperscript{2}}\quad
      \mbox{Steven Hoi\textsuperscript{2}}\quad
      \mbox{Hongyang Li\textsuperscript{1}}\endgraf}
      \vspace{0.6em}
      {\small\normalfont
      \mbox{\textsuperscript{1}OpenDriveLab at The University of Hong Kong}\quad
      \mbox{\textsuperscript{2}Alibaba Group}\quad
      \mbox{\textsuperscript{3}Shanghai Innovation Institute}\quad
      \mbox{\textsuperscript{4}Fudan University}\quad
      \mbox{\textsuperscript{5}Beihang University}\quad
      \mbox{\textsuperscript{6}Sichuan University}\quad
      \mbox{\textsuperscript{7}Archon Robotics}\quad
      \endgraf}
      \vspace{0.6em}
      {\centering\normalfont\normalsize
      {\hypersetup{urlcolor=qwenpurple}\texttt{\url{https://\demowebsite}}}\endgraf}}}

\begin{document}

\maketitle
\begingroup
  \renewcommand{\thefootnote}{\fnsymbol{footnote}}
  \footnotetext[2]{Work done during an internship at Alibaba Token Hub (ATH), Alibaba Group.}
\endgroup
\fancyhead{}
  \fancyhead[L]{Preprint}
  \renewcommand{\headrulewidth}{0.4pt}

\begin{center}
  \vspace{-2.7em}
  \includegraphics[width=0.99\textwidth]{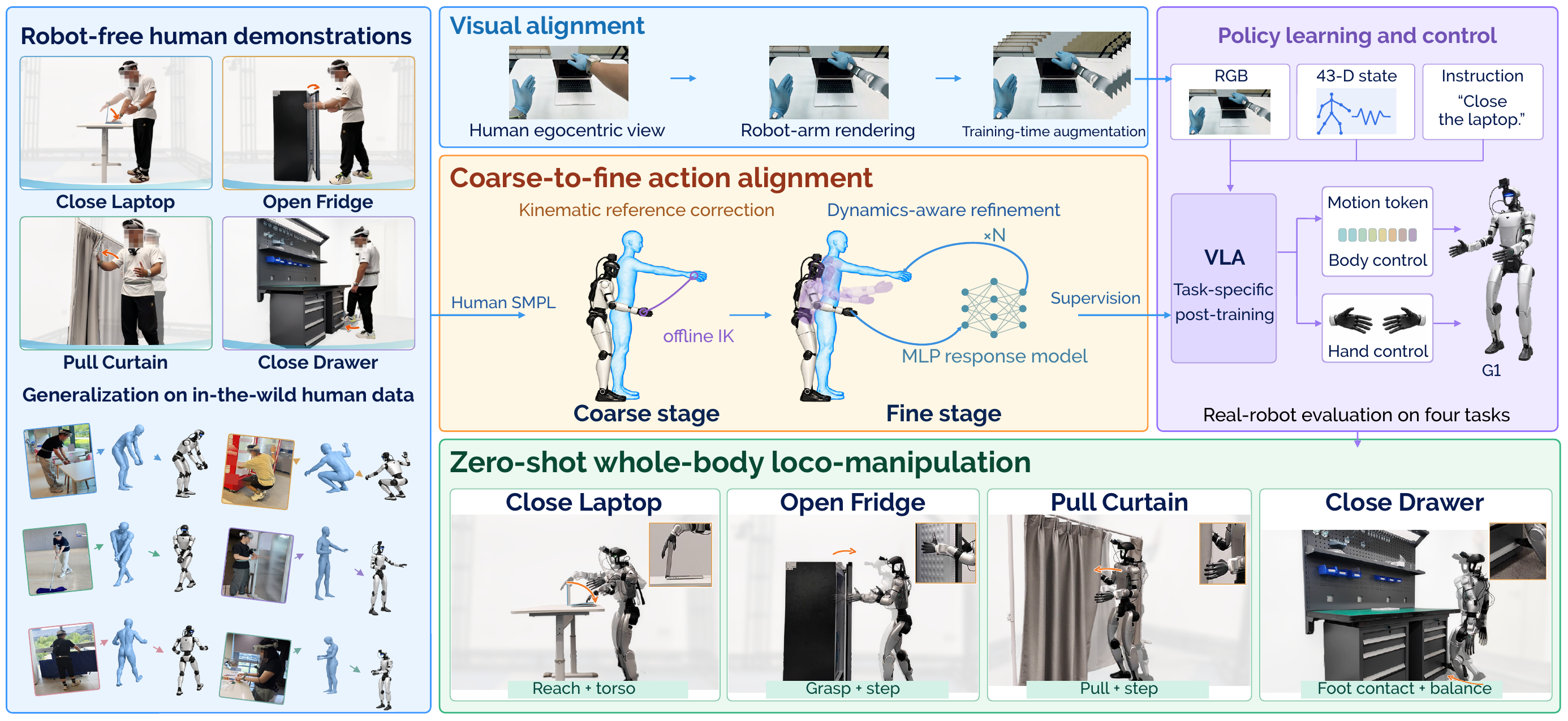}
  \captionof{figure}{Introducing \textbf{EgoHumanoid-V2}, the first egocentric
  human-to-humanoid skill transfer framework for coordinated whole-body
  loco-manipulation.
  Robot teleoperation is costly and hard to scale across environments,
  while human demonstrations capture diverse scenes and coordinated whole-body
  skills. Action and visual alignment bridge the embodiment gap for VLA training.
  Four real-world tasks demonstrate zero-shot skill transfer without
  target-task robot demonstrations.}
  \label{fig:teaser}
\end{center}

\begin{abstract}
Human demonstrations capture diverse scenes and rich whole-body skills
without requiring robot teleoperation.
Prior work on egocentric transfer has emphasized scene generalization in
loco-manipulation under decoupled control, leaving direct transfer
of coordinated whole-body skills less explored.
We present \textbf{EgoHumanoid-V2}, the first egocentric human-to-humanoid
skill transfer framework for coordinated whole-body loco-manipulation.
At its core, coarse-to-fine action alignment combines kinematic
reference correction with dynamics-aware refinement. It improves end-effector
pose accuracy while preserving whole-body coordination.
We also use robot-arm rendering and training-time image augmentation to reduce
the visual embodiment gap and improve viewpoint robustness.
On four real-world tasks, vision-language-action (VLA) policies trained on
aligned human data show zero-shot skill transfer without target-task robot
demonstrations. Task scores are comparable to those of policies trained on
teleoperation data at a lower collection cost. These results support human data
as direct skill supervision.
\end{abstract}

\section{Introduction}
\label{sec:introduction}

Human demonstrations offer rich manipulation experience across diverse
scenes without robot teleoperation \citep{grauman2022ego4d,grauman2024egoexo4d}.
Yet the embodiment gap
makes it challenging to exploit human motion for robot policy learning.
Some approaches learn loco-manipulation priors from videos without action
labels
\citep{ye2024lapa,univla,jiang2025wholebodyvla,ye2026dreamzero}.
This bypasses explicit action alignment, but may underutilize human motion
\citep{lin2026cotraininglbm}.
Recent work captures human hand trajectories with the Universal
Manipulation Interface (UMI) \citep{chi2024umi,liu2026rdt2} or VR motion
capture \citep{yuan2025motiontrans,shi2026egohumanoid}.
Relative end-effector pose changes provide a shared
human–robot action space \citep{kareer2025emergence,yuan2025motiontrans}.
However, these methods primarily transfer bimanual manipulation and navigation,
rather than coordinated whole-body loco-manipulation skills.

Humans naturally coordinate their arms, torso, and legs to extend their
workspace and manipulate efficiently. Transferring these coordinated skills,
however, requires more than matching end-effector trajectories. Whole-body
tracking policies enable such coordination through joint whole-body
control \citep{luo2025sonic,pan2025ams}.
Unlike decoupled controllers that support arm inverse kinematics (IK)
alongside separate balance and locomotion control \citep{homie,amo},
these policies jointly generate commands from full-body feedback.
However, tracking human references alone can leave substantial
end-effector error (Table~\ref{tab:action-transfer}).
A straightforward approach is to correct the tracker's joint commands
using online IK. Yet these corrections override part of the coordinated
output without adjusting the remaining commands, potentially disrupting
balance. Our implementation and evaluation confirm this risk
(Appendix~\ref{sec:online-ik-instability}).
The challenge is therefore to improve interaction accuracy while
preserving whole-body coordination.

We present \textbf{EgoHumanoid-V2}, the first egocentric human-to-humanoid
skill transfer framework for coordinated whole-body loco-manipulation
(Figure~\ref{fig:teaser}). Building on EgoHumanoid's exploration of egocentric
human-to-humanoid transfer \citep{shi2026egohumanoid}, we extend the scope from
loco-manipulation under decoupled control to coordinated whole-body skills.
At its core is offline, coarse-to-fine action alignment that accounts for
the tracking policy's execution. In the coarse stage,
kinematic reference correction uses offline IK to align motion references with
human hand or foot targets, reducing geometric mismatch before tracking.
In the fine stage, dynamics-aware refinement combines rollout feedback with
a learned local response model to iteratively improve end-effector tracking.
Together, these stages improve manipulation accuracy while preserving whole-body
coordination.
We also use robot-arm rendering to reduce the visual embodiment gap and
training-time image augmentation to improve viewpoint robustness.
The resulting observations and actions provide paired supervision for
vision-language-action (VLA) post-training.

On four real-world tasks, policies trained on 200 human demonstrations per
task achieve a mean score of 51.67\% without target-task robot demonstrations.
On two tasks, human collection requires 76.36\% less labor per demonstration
than teleoperation. Ablations examine action alignment through tracking
accuracy and motion quality, and visual alignment through policy performance.

Our contributions are:
\textbf{(a)} The first egocentric human-to-humanoid skill transfer framework
for coordinated whole-body loco-manipulation.
\textbf{(b)} Coarse-to-fine action alignment through
kinematic reference correction and dynamics-aware refinement.
\textbf{(c)} Evidence of zero-shot skill transfer on four real-world tasks,
with analyses of accuracy, policy scores, and collection cost.

\section{Related Work}
\label{sec:related-work}

\subsection{Whole-Body Control for Loco-Manipulation}
Humanoid whole-body control supports imitation \citep{exbody2,humanplus}
and teleoperation \citep{h2o,omnih2o,twist}. For loco-manipulation, decoupled
controllers combine arm joint targets or IK with adaptive lower-body control
\citep{homie,amo}. Whole-body tracking policies instead use motion references
and full-body feedback for joint control \citep{gmt,pan2025ams}, supporting
diverse motions \citep{luo2025sonic,qi2026humanoidgpt} and precise spatial
tracking \citep{xu2026glori}.
Retargeting improves reference quality \citep{joao2025gmr}, interaction
geometry \citep{yang2025omniretarget}, and dynamic feasibility \citep{pan2025spider}.
Execution errors nevertheless remain. Residual learning addresses dynamics
mismatch \citep{asap} and task interactions \citep{zhao2025resmimic}.
We instead correct references and refine execution offline under SONIC,
keeping the controller fixed across tasks.

\subsection{Learning from Human Demonstrations}
Human demonstrations support robot learning through visual representations
\citep{nair2022r3m,ma2022vip,mpi}, predictive pretraining \citep{wu2023gr1},
and motion priors from latent action models
\citep{ye2024lapa,univla,jiang2025wholebodyvla} or world action models
\citep{ye2026dreamzero}. Captured human trajectories provide direct action
supervision \citep{hoque2025egodex} through UMI-based interfaces
\citep{chi2024umi,zhaxizhuoma2025fastumi,xu2025dexumi,liu2026rdt2}
or VR capture \citep{yuan2025motiontrans}. Wearable hand capture enables
human--robot co-training \citep{kareer2025egomimic,qiu2025hat,tao2025dexwild},
with domain adaptation improving transfer \citep{punamiya2025egobridge}.
Recent egocentric systems study scaling \citep{zheng2026egoscale} and
humanoid control \citep{wei2026psi0,yang2026zerowbc}.
Loco-manipulation transfer extends to wheeled robots \citep{xu2026hommi}
and humanoids with decoupled control \citep{shi2026egohumanoid,zhong2025humanoidexo}.
Scene reconstruction supports contextual whole-body imitation \citep{videomimic},
while HuMI and BifrostUMI transfer coordinated manipulation
\citep{nai2026humi,wang2026bifrostumi}.
We instead refine action supervision offline under a fixed tracker, without
task-specific training or fine-tuning.

\section{Method}
\label{sec:method}

Figure~\ref{fig:alignment} illustrates our action and visual alignment pipelines.

\subsection{Problem Setup}
\label{sec:problem-setup}
For each task $\tau$, human demonstrations contain synchronized egocentric images
$I$, body motion $x$, and hand signals $h$, together with an instruction $\ell$.
We convert them into robot observations and actions to post-train a VLA policy
$\pi_\theta^\tau$ without target-task robot demonstrations. The policy predicts
whole-body and hand actions from images, proprioceptive states, and instructions.

\Needspace{6\baselineskip}
\subsection{Data Collection and Policy Learning}
\label{sec:data-policy-pipeline}
We use a wearable system based on EgoHumanoid \citep{shi2026egohumanoid} to
capture human demonstrations without robot hardware
(Figure~\ref{fig:hardware-setup}).

\Needspace{17\baselineskip}
\begingroup
\setlength{\intextsep}{2pt}
\begin{wrapfigure}{r}{0.5\textwidth}
  \vspace{-4pt}
  \centering
  \captionsetup{skip=2pt}
  \includegraphics[width=\linewidth]{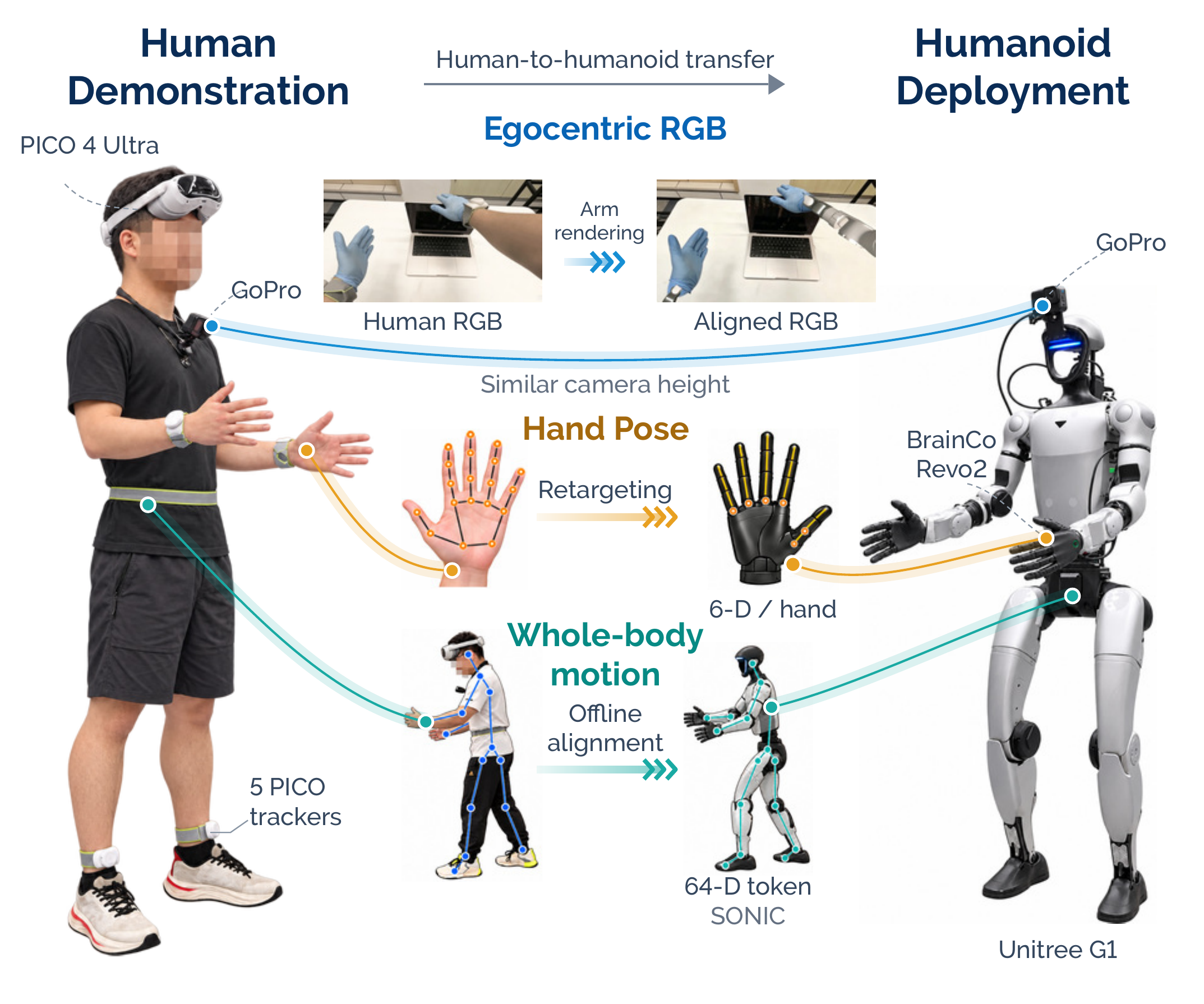}
\caption{Hardware for human demonstration capture and humanoid deployment.}
  \label{fig:hardware-setup}
  \vspace{-2pt}
\end{wrapfigure}
\paragraph{Human demonstration capture.}
A neck-mounted GoPro captures egocentric RGB images, while the PICO system
captures body motion and hand poses. These streams are synchronized at 50~Hz.
For deployment, we use a Unitree G1 with a head-mounted GoPro and BrainCo Revo2
hands. The human and robot cameras are placed at similar heights to reduce the viewpoint gap.

\paragraph{Training data construction.}
Offline action alignment converts human motion into SONIC tokens
\citep{luo2025sonic}. Controller rollouts provide humanoid states for policy
inputs and robot-arm rendering. The adapted images and aligned actions form
training pairs, with image augmentation applied during training.

\par\endgroup

\paragraph{Policy learning and deployment.}
Using the aligned human demonstrations, we post-train one $\pi_{0.5}$ policy
\citep{intelligence2025pi05} per task without target-task robot demonstrations.
The policy predicts action sequences from egocentric images, humanoid states,
and language instructions, with a common interface for training and deployment:
\begin{equation}
o_t=(\widetilde I_t,q_t,\ell), \qquad
a_t=(z_t,h_t^{\mathrm L},h_t^{\mathrm R}),
\label{eq:runtime-interface}
\end{equation}
where $\widetilde I_t$ denotes the adapted human image during training or the
robot camera image at deployment, $q_t$ is the 43-D G1 state,
$z_t\in\mathbb{R}^{64}$ is a SONIC motion token, and
$h_t^{\mathrm L/R}\in\mathbb{R}^{6}$ are Revo2 hand commands.
Controller rollouts supply $q_t$ in the robot's state space during training.
At deployment, SONIC executes motion tokens and the Revo2 driver executes
hand commands. The same SONIC checkpoint serves alignment and deployment
across all four tasks.

\subsection{Action Alignment}
\label{sec:action-alignment}
Action alignment (Figure~\ref{fig:alignment}(a)) targets accurate interaction
trajectories while preserving coordinated control. Starting from a controller rollout ($H$),
the coarse stage corrects reference geometry ($K$), and the fine stage iteratively
refines motion tokens using execution feedback ($D$).

\paragraph{Motion initialization.}
We obtain an initial robot motion by executing human body motion through SONIC
in simulation, retaining the resulting joint trajectory $b^0_{1:T}$ and motion
tokens, where $b_t\in\mathbb{R}^{29}$. We denote this initialization by $H$.
Other motion references are also compatible with the framework; we compare
their accuracy and motion quality in Table~\ref{tab:initialization}.

\paragraph{Target construction.}
We express motion in pelvis-centered frames with vertical axes fixed upright
and horizontal axes following pelvis yaw. Human palm poses define the targets,
with heights adjusted by the per-frame human--robot pelvis-height difference
from the initial rollout. Targets remain fixed throughout refinement.
Palm and sole target definitions appear in
Appendix~\ref{sec:alignment-implementation}.

\subsubsection{Kinematic Reference Correction (\texorpdfstring{$K$}{K})}
The coarse stage uses constrained offline IK to bring the initial robot
reference closer to the human targets. The optimization reduces end-effector
pose error while limiting changes to the original motion and maintaining
temporal continuity. We encode the corrected reference $b_{1:T}^{\mathrm{ref}}$
using SONIC's robot motion encoder $\mathcal{E}_r$ \citep{luo2025sonic}:
\begin{equation}
z_{1:T}^{0}=\mathcal{E}_r(b_{1:T}^{\mathrm{ref}}).
\label{eq:reference-alignment}
\end{equation}
Token replay measures the residual tracking errors corrected in the fine stage
(Appendix~\ref{sec:alignment-implementation}).

\subsubsection{Dynamics-Aware Refinement (\texorpdfstring{$D$}{D})}
The fine stage uses rollout feedback and an MLP response model
$\mathcal{R}_{\psi}$. Paired perturbed and unperturbed rollouts train
it to predict execution changes from kinematic changes
(Appendix~\ref{sec:response-training}).

At iteration $k$, a rollout of $z^k$ provides the controller observation history
$s^k$ and end-effector residual $r^k$: position errors and, for hand targets,
relative rotation vectors from executed poses to fixed targets.
Holding the rollout context fixed, SONIC's control decoder $\mathcal{D}_c$
maps candidate tokens to joint commands. We predict their effect on execution as
\begin{equation}
\begin{aligned}
\Delta u(z)&=\mathcal{D}_c(z,s^k)-\mathcal{D}_c(z^k,s^k),\\
\widehat{\Delta e}(z)&=\mathcal{R}_\psi\!\left(
J^k\Delta u(z)\right).
\end{aligned}
\label{eq:response-prediction}
\end{equation}
Here, $J^k$ maps joint-command changes to kinematic end-effector changes.
We optimize the tokens so that the predicted response
compensates for the current residual:
\begin{equation}
\mathcal{L}_{\mathrm{track}}(z)=
\|\widehat{\Delta e}(z)-r^k\|^2.
\label{eq:closed-loop-refinement}
\end{equation}
Gradients pass through the fixed response model and control decoder, not the
simulator. Each updated sequence is evaluated in a fresh rollout, which supplies
the residual and context for the next iteration.
We select a validated sequence from the optimization history for policy
training. The selected output, $D_s$, may come from an earlier round,
including $H$ or $H+K$. The full formulation adds temporal history, state
and token conditioning, task weights, and regularization; selection criteria
and implementation details appear in
Appendices~\ref{sec:alignment-implementation}--\ref{sec:data-acceptance}.

\subsection{Visual Alignment}
\label{sec:visual-alignment}
\begin{figure*}[t]
  \centering
  \includegraphics[width=0.95\textwidth]{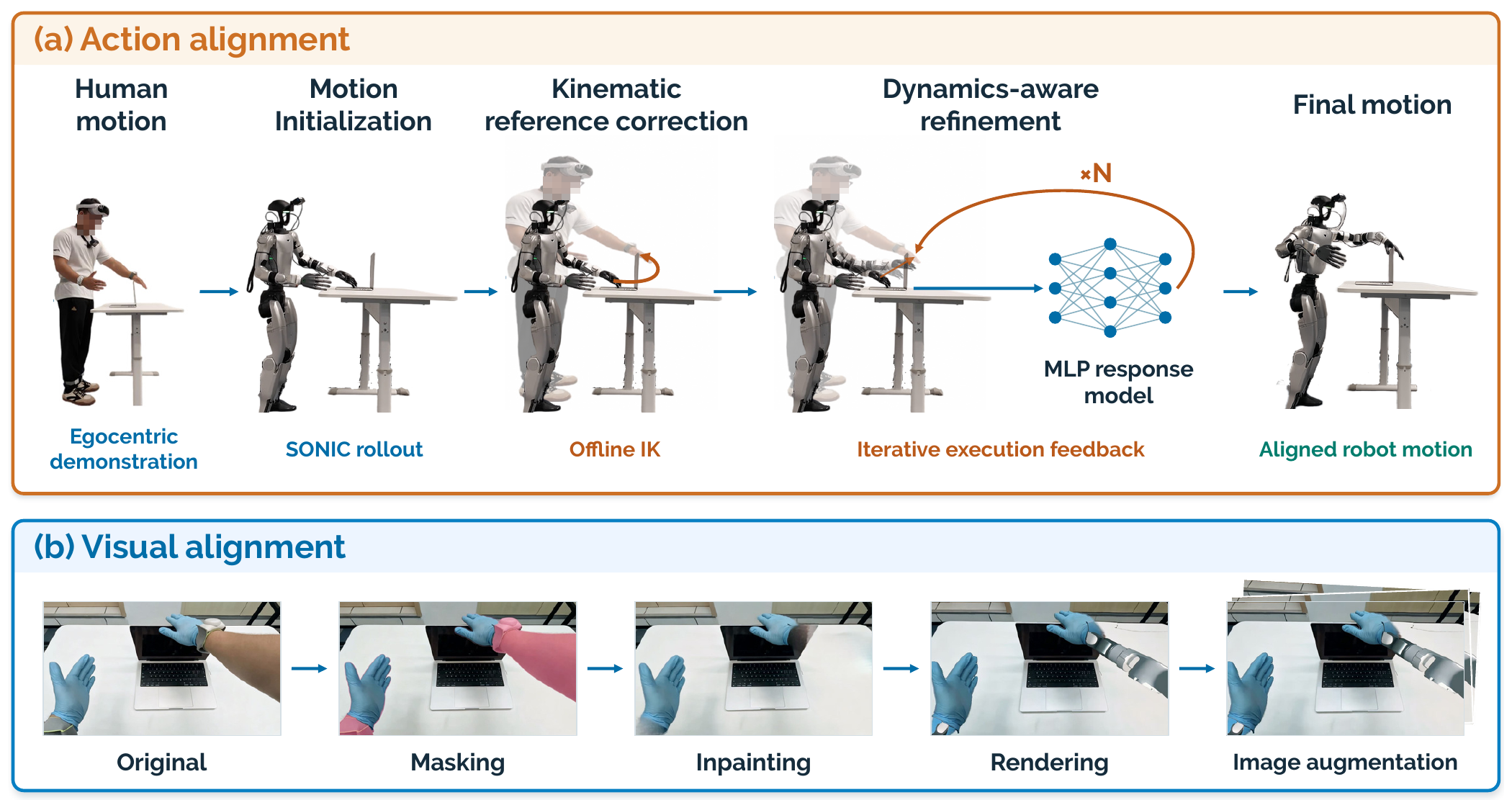}
  \caption{Action and visual alignment pipelines.
  (a) Human motion initializes a robot trajectory through a SONIC rollout ($H$).
  Offline IK corrects the reference geometry ($K$), and dynamics-aware
  refinement ($D$) uses an MLP response model and iterative execution feedback
  to produce aligned robot motion.
  (b) Human forearms and wrist trackers are masked and inpainted, then replaced
  with rendered robot arms. Image augmentation produces varied training views.}
  \label{fig:alignment}
\end{figure*}

Action alignment alone does not close the visual gap between human
demonstrations and robot deployment (Table~\ref{tab:visual-ablation}).
We address differences in arm appearance and camera viewpoint through
robot-arm rendering and training-time image augmentation
(Figure~\ref{fig:alignment}(b)).

\paragraph{Robot-arm rendering.}
We adapt arm appearance~\citep{chen2024mirage,chen2024rovi}.
Matching blue gloves on humans during collection and robots during deployment
simplify segmentation and visual alignment.
We use SAM2~\citep{ravi2024sam2} to segment human forearms and wrist-mounted
PICO trackers, then fill the masked regions with DiffuEraser~\citep{li2025diffueraser}.
We render G1 arms from aligned joint positions and adjust their translation,
rotation, and uniform scale to match human wrists and forearms before
compositing onto the inpainted backgrounds.

\paragraph{Image augmentation.}
Differences in movement amplitude and speed can alter camera viewpoints
despite matching the initial camera height and angle. Random resized crops,
small rotations, and photometric jitter improve robustness to these variations.
At inference, the policy uses robot images directly
(Appendix~\ref{sec:visual-training-details}). Table~\ref{tab:visual-ablation}
evaluates rendering and augmentation separately and together.

\Needspace{5\baselineskip}
\section{Experiments}
\label{sec:experiments}

We evaluate skill transfer on a real humanoid and use simulation to examine
the accuracy and motion quality of action alignment. We ask four questions:
\par
\begingroup
\setlength{\parskip}{3pt}
\noindent\textbf{Q1:} Can human demonstrations teach whole-body loco-manipulation skills?
\par\noindent\textbf{Q2:} How do human and robot data compare in performance and cost?
\par\noindent\textbf{Q3:} How does action alignment improve accuracy?
\par\noindent\textbf{Q4:} How does visual alignment improve transfer?
\par\endgroup

\subsection{Experimental Setup}
\label{sec:experimental-setup}
We evaluate four coordinated whole-body manipulation tasks on Unitree G1,
covering hand and foot interactions with posture adjustments or stepping
(Figure~\ref{fig:task-scaling-overview}).

\noindent\mbox{(1) \texttt{Close Laptop.}} The robot reaches above the laptop
with its right hand and presses the lid shut, testing reaching accuracy and
controlled downward contact until the lid is closed.
\mbox{(2) \texttt{Open Fridge.}} The robot hooks its left hand
into the door gap and pulls the door open, requiring precise hand placement
and sustained contact as the robot adjusts its posture.
\mbox{(3) \texttt{Pull Curtain.}} The robot grasps the curtain
with its right hand and pulls it open by one meter, combining grasp
maintenance with arm and body motion over an extended pulling distance.
\mbox{(4) \texttt{Close Drawer.}} The robot lifts its right foot
and pushes the open drawer fully closed with its toe, requiring accurate
foot placement while maintaining balance.

For each task, we post-train a separate $\pi_{0.5}$ policy for 30,000 steps
with batch size 256, using the same pretrained checkpoint and training
settings (Appendix~\ref{sec:post-training-config}). Unless otherwise specified,
human demonstrations undergo the full $H+K+D_s$ action-alignment pipeline,
robot-arm rendering, and image augmentation; Close Drawer omits arm
rendering. Human-data policies use no target-task robot demonstrations during
post-training and are deployed directly on the robot.

Each policy is evaluated over 20 real-world trials with three sequential
subtasks, each scored as success or failure. Averaging over subtasks and trials
gives a task score that captures partial completion
(Appendix~\ref{sec:task-success}). Offline endpoint-error and motion-quality
comparisons use a fixed random subset of 25 training trajectories per task
(Appendices~\ref{sec:offline-metrics}--\ref{sec:initialization-details}).
Cross-task means weight tasks equally.

\subsection{Q1: Skill Transfer from Human Demonstrations}
\label{sec:zero-shot-evaluation}
\begin{figure*}[t]
  \centering
  \includegraphics[width=0.99\textwidth]{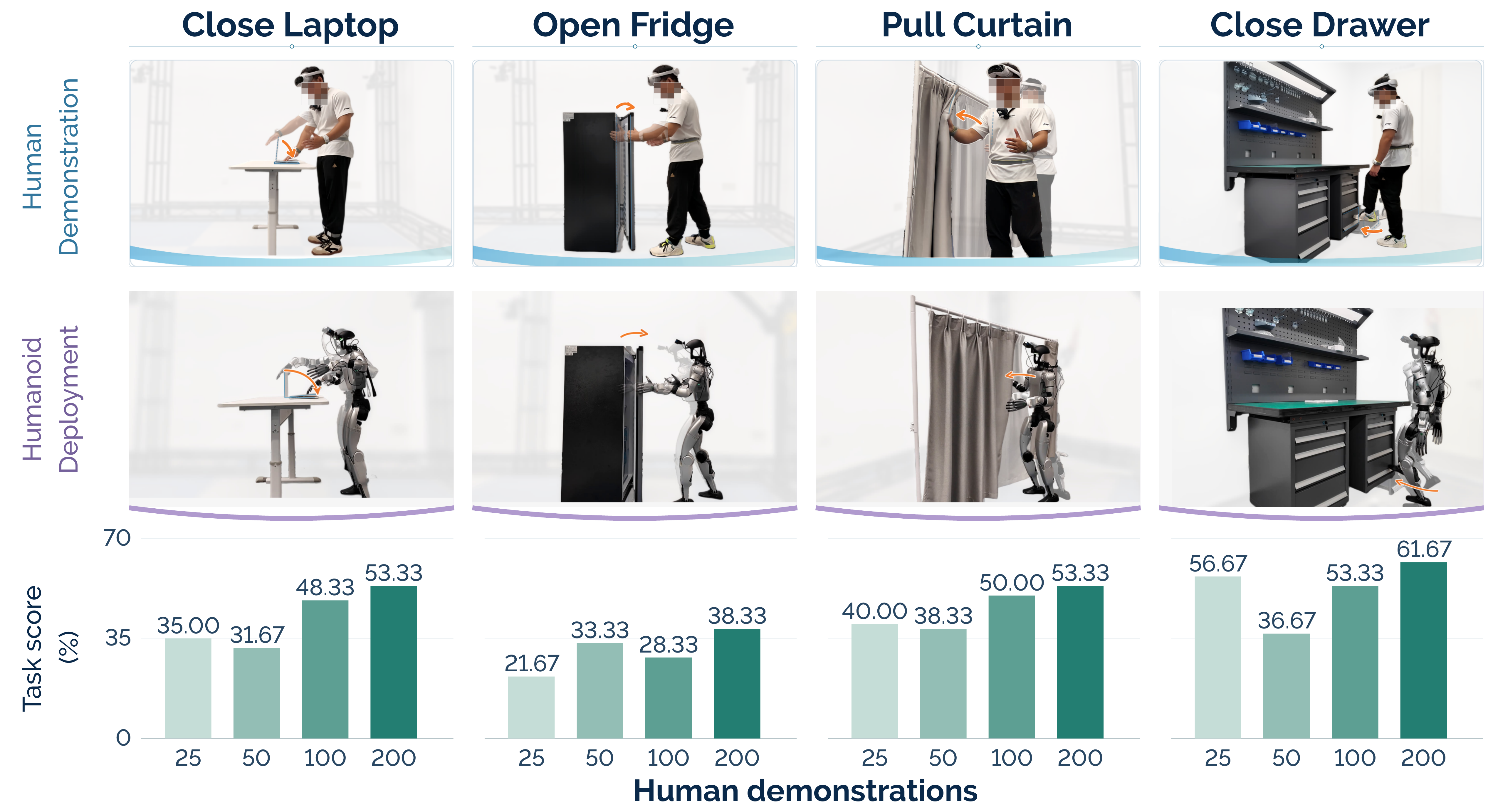}
  \caption{\textbf{Q1: Skill transfer on four real-world tasks.}
  Each task column shows human demonstrations (top), humanoid motion illustrations
  (middle), and task scores with 25, 50, 100, or 200 human demonstrations (bottom).
  Policies use no target-task robot demonstrations.}
  \label{fig:task-scaling-overview}
  \label{fig:tasks}
  \label{fig:scaling}
\end{figure*}

Figure~\ref{fig:task-scaling-overview} evaluates transfer with 25, 50, 100,
and 200 human demonstrations per task.

With 200 human demonstrations per task, scores range from 38.33\% on Open
Fridge to 61.67\% on Close Drawer, with a four-task mean of 51.67\%.
These results show that human demonstrations supervise both arm and foot
manipulation through the same whole-body controller. Increasing the budget
from 25 to 200 raises the mean from 38.33\% to 51.67\%, suggesting that
additional human demonstrations improve transfer to robot deployment.

\Needspace{4\baselineskip}
\subsection{Q2: Human vs. Robot Demonstrations}
\label{sec:data-efficiency}
\begin{figure}[!htbp]
  \centering
  \includegraphics[width=0.82\linewidth]{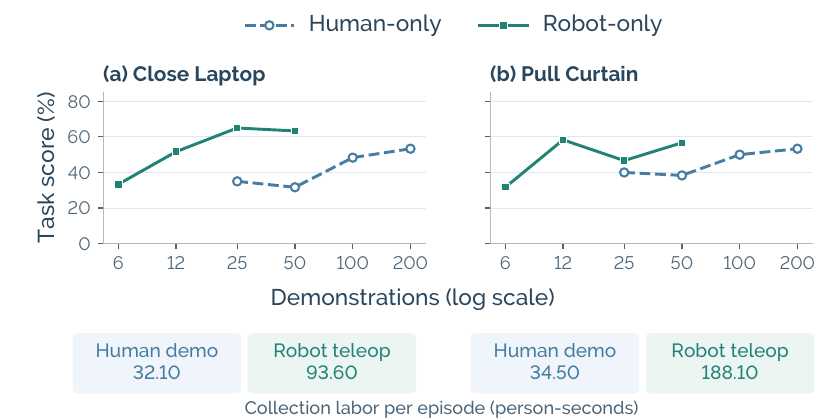}
  \caption{\textbf{Task scores and collection labor with human or robot demonstrations.}
  Each point represents an independently trained policy evaluated over 20 trials.
  Human-only results reuse Figure~\ref{fig:task-scaling-overview}.
  Cost strips include capture, setup, calibration, troubleshooting, and resets.}
  \label{fig:data-efficiency}
\end{figure}
Figure~\ref{fig:data-efficiency} compares human and robot demonstrations on
Close Laptop and Pull Curtain under the same training and evaluation settings.
Robot-data policies use 6, 12, 25, or 50 demonstrations. With 200 human
demonstrations, both tasks reach 53.33\%, compared with 63.33\% and 56.67\%,
respectively, from 50 robot demonstrations. Human data approach robot-data
performance but require more episodes, motivating a comparison of collection labor.

We measure collection labor in person-seconds, summing time across operators
for capture, setup, calibration, troubleshooting, and scene resets. Shared
overhead is amortized over episodes. Human data collection uses one operator,
whereas our robot data collection uses two: one for teleoperation and one for
scene and robot resets. Collecting a robot demonstration requires
2.92$\times$ as much labor as a human demonstration on Close Laptop and
5.45$\times$ on Pull Curtain. On average, human collection reduces labor
per episode from 140.85 to 33.30 person-seconds, a 76.36\% reduction.

On Close Laptop, 100 human demonstrations achieve 48.33\% versus 63.33\%
from 50 robot demonstrations, using 31.41\% less collection labor. On Pull
Curtain, 200 human demonstrations reach 53.33\% versus 56.67\% from 50 robot
demonstrations, using 26.63\% less labor.

\subsection{Q3: Action Alignment for Improved Accuracy}
\label{sec:action-ablation}
\paragraph{Why correct actions offline?}
We first examine whether online IK can improve endpoint accuracy during
whole-body tracking. All three tested implementations produce unstable
SONIC rollouts, including falls, across the four tasks, so we do not proceed
to policy evaluation with these variants. Their designs and failure sequences
are reported in Appendix~\ref{sec:online-ik-instability}.

\paragraph{Kinematic reference correction and dynamics-aware refinement.}
Table~\ref{tab:action-transfer} compares $H$, $H+K$, and the selected
$H+K+D_s$ output under the same controller. Position error covers all four
tasks; orientation error covers the three palm-interaction tasks. Drawer
closing uses sole position only.

\begin{table}[!htb]
  \centering
  \caption{\textbf{Q3: Effect of action alignment.}
  Top: position (cm) / orientation ($^\circ$) errors in simulation ($\downarrow$).
  Bottom: real-world task scores (\%, $\uparrow$) with 200 demonstrations per
  task and identical visual preprocessing. Means are computed across tasks;
  Close Drawer uses sole position only and is excluded from orientation averages.
  Bold/underline mark best/second best.
  Selection follows Appendix~\ref{sec:stage-selection}.}
  \label{tab:action-transfer}
  \papertablestyle
  \setlength{\tabcolsep}{1.7pt}
  \begin{tabular}{@{}lccccc@{}}
    \toprule
    \multicolumn{6}{c}{\cellcolor{papertealtint}\textbf{End-effector Tracking Error ($\downarrow$)}} \\
    \multirow{2}{*}{Alignment} & Close Laptop & Open Fridge & Pull Curtain & Close Drawer & Mean \\
    & \poseerrorheader & \poseerrorheader & \poseerrorheader
      & Pos. (cm) & \poseerrorheader \\
    \midrule
    $H$ & 14.08 / 29.43 & 13.27 / 19.37 & 10.31 / 14.86 & 7.17 & 11.21 / 21.22 \\
    $H+K$ & \underline{4.22} / \underline{13.84} & \underline{4.24} / \underline{14.72} & \underline{3.13} / \underline{12.74} & \underline{6.22} & \underline{4.45} / \underline{13.77} \\
    \rowcolor{papertealtint}
    $H+K+D_s$ & \textbf{3.04} / \textbf{9.50} & \textbf{2.76} / \textbf{5.77} & \textbf{2.02} / \textbf{4.85} & \textbf{5.80} & \textbf{3.41} / \textbf{6.71} \\
    \midrule
    \multicolumn{6}{c}{\cellcolor{paperbluetint}\textbf{Real-World Task Score (\%, $\uparrow$)}} \\
    \midrule
    $H$ & 0.00 & 0.00 & 28.33 & 56.67 & 21.25 \\
    $H+K$ & \underline{46.67} & \underline{31.67} & \underline{51.67}
      & \underline{58.33} & \underline{47.08} \\
    \rowcolor{paperbluetint}
    $H+K+D_s$ & \textbf{53.33} & \textbf{38.33} & \textbf{53.33}
      & \textbf{61.67} & \textbf{51.67} \\
    \bottomrule
  \end{tabular}
\end{table}

Mean position error decreases from 11.21~cm at $H$ to 4.45~cm after $K$ and
3.41~cm at $H+K+D_s$; orientation error falls from 21.22$^\circ$ to
13.77$^\circ$ and 6.71$^\circ$. Thus, $K$ removes much of the initial position
mismatch, while $D$ further improves execution accuracy, especially in
orientation. With 200 demonstrations per task, mean
policy score likewise rises from 21.25\% to 47.08\% and 51.67\%.
The full pipeline improves policy scores over $H+K$ on every task.
An expanded simulation study on 112 trajectories across 50 tasks shows
the same stage-wise trend: position error falls from 17.03 to 5.22 to
3.83~cm (Appendix~\ref{sec:expanded-task-benchmark},
Table~\ref{tab:expanded-stage-comparison}).

\paragraph{Motion initialization.}
Table~\ref{tab:initialization} compares three motion initialization strategies: a SONIC
rollout, GMR retargeting \citep{joao2025gmr}, and whole-body IK adapted from
HuMI \citep{nai2026humi}. Each passes through the same subsequent $K$ stage,
eight refinement rounds, and selection procedure.

\begin{table}[!htb]
  \centering
  \caption{\textbf{Effect of motion initialization.}
  Task-averaged endpoint error at $D_s$ and whole-body motion quality.
  Pelvis drop measures maximum height loss; tilt, maximum base inclination;
  joint step, maximum frame-to-frame joint-angle change; extra yaw, heading
  deviation from the human reference. All variants share $K$, eight $D$ rounds,
  and selection. Drawer orientation is excluded. Lower is better;
  bold/underline mark best/second best.
  Appendix~\ref{sec:motion-quality-metrics} defines the metrics;
  Tables~\ref{tab:initialization-taskwise} and~\ref{tab:motion-quality-taskwise}
  report results for each task and initialization.}
  \label{tab:initialization}
  \papertablestyle
  \begin{tabular}{@{}lcc@{\hspace{9pt}}cccc@{}}
    \toprule
    & \multicolumn{2}{c}{\cellcolor{papertealtint}\textbf{Endpoint Error}}
    & \multicolumn{4}{c}{\cellcolor{paperbluetint}\textbf{Whole-Body Motion Quality}} \\
    \cmidrule(lr){2-3} \cmidrule(lr){4-7}
    Motion initialization & Pos. (cm) & Ori. ($^\circ$) & Pelvis drop (cm)
      & Tilt ($^\circ$) & Joint step ($^\circ$) & Extra yaw ($^\circ$) \\
    \midrule
    SONIC Rollout & \textbf{3.41} & 6.71 & \textbf{3.28}
      & \textbf{11.14} & \textbf{5.94} & \textbf{7.18} \\
    Retargeting & \underline{3.42} & \textbf{5.32} & \underline{3.30}
      & \underline{11.87} & \underline{6.51} & \underline{11.45} \\
    Whole-Body IK & 3.82 & \underline{6.35} & 5.25
      & 12.77 & 7.75 & 23.67 \\
    \bottomrule
  \end{tabular}
\end{table}

SONIC rollout and retargeting achieve similar position errors (3.41 versus
3.42~cm). Retargeting gives lower orientation error (5.32$^\circ$ versus
6.71$^\circ$), while SONIC rollout reduces extra yaw from 11.45$^\circ$ to
7.18$^\circ$ and improves the other motion-quality metrics. This balance
of accuracy and motion quality motivates our use of SONIC rollout for motion initialization.

\begin{figure}[!htb]
  \centering
  \includegraphics[width=\linewidth]{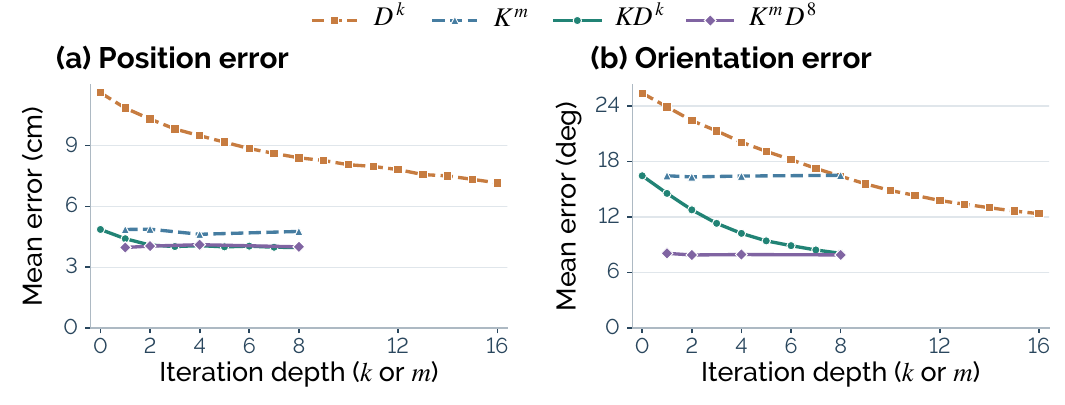}
  \caption{\textbf{Complementarity and depth of $K$ and $D$.}
  Task-macro averages of raw stage outputs for $D^k$, repeated $K^m$, $KD^k$,
  and $K^mD^8$. Position averages four tasks; orientation averages the three
  hand tasks. Protocol and processing costs appear in
  Appendix~\ref{sec:refinement-depth-details}.}
  \label{fig:kd-iterations}
\end{figure}

\paragraph{Roles of $K$ and $D$.}
Figure~\ref{fig:kd-iterations} supports one $K$ round followed by multiple $D$
rounds. Repeating $K$ eight times changes position error only from 4.86 to
4.77~cm. Instead, adding eight $D$ rounds after one $K$ reduces position
error to 3.97~cm and orientation error from 16.46$^\circ$ to 8.08$^\circ$.
Eight $D$ rounds alone leave errors of 8.40~cm and 16.40$^\circ$.
Thus, $K$ provides an accurate reference, while repeated $D$ updates further
reduce execution error; repeating $K$ offers little additional benefit.
The 50-task experiments likewise support one $K$ round followed by
multiple $D$ rounds
(Figure~\ref{fig:benchmark-refinement-depth}).

\paragraph{Learning the execution response.}
Table~\ref{tab:response-model} compares the learned MLP with a unit response
that equates kinematic and executed endpoint changes. Under identical settings,
the MLP improves all four tasks: arm-task errors fall by 0.13--0.17~cm and
0.11--0.24$^\circ$, while Close Drawer improves from 9.18 to 5.80~cm.
The larger drawer gain suggests that modeling execution is especially useful
when foot corrections require coordinated balance adjustments.
On 50 tasks, raw $KD^8$ errors also improve from 4.25 to 4.10~cm and
9.39$^\circ$ to 9.21$^\circ$ (Table~\ref{tab:expanded-response-model};
training details in Appendix~\ref{sec:response-training}).

\begin{table}[!htb]
  \centering
  \caption{\textbf{MLP response versus unit response.} Position (cm) /
  orientation ($^\circ$) errors of historically selected outputs within eight
  refinement rounds, with shared inputs and evaluation trajectories.
  Means weight tasks equally; orientation excludes drawer. Bold marks lower
  errors (Appendix~\ref{sec:response-ablation}).}
  \label{tab:response-model}
  \papertablestyle
  \setlength{\tabcolsep}{1.7pt}
  \begin{tabular}{@{}lccccc@{}}
    \toprule
    Response & Close Laptop & Open Fridge & Pull Curtain & Close Drawer & Mean \\
    & \poseerrorheader & \poseerrorheader & \poseerrorheader
      & Pos. (cm) & \poseerrorheader \\
    \midrule
    Unit response & 3.18 / 9.74
      & 2.89 / 5.88
      & 2.19 / 5.06
      & 9.18 & 4.36 / 6.89 \\
    \rowcolor{papertealtint}
    MLP response & \textbf{3.04} / \textbf{9.50}
      & \textbf{2.76} / \textbf{5.77}
      & \textbf{2.02} / \textbf{4.85}
      & \textbf{5.80} & \textbf{3.41} / \textbf{6.71} \\
    \bottomrule
  \end{tabular}
\end{table}

\subsection{Q4: Visual Alignment for Easier Transfer}
\label{sec:visual-ablation}
We evaluate robot-arm rendering and image augmentation separately and
together, using 200 human demonstrations per task with identical action
labels and training settings (Table~\ref{tab:visual-ablation}). The baseline
uses human images with standard model preprocessing only.

\begin{table}[!htb]
  \centering
  \caption{\textbf{Q4: Visual-alignment ablation.} Task scores with 200 human
  demonstrations per task. Means average the available tasks; robot-arm
  rendering is not evaluated for foot-operated Close Drawer.
  Bold/underline mark best/second best;
  shading indicates the full pipeline.}
  \label{tab:visual-ablation}
  \papertablestyle
  \begin{tabular}{@{}cc*{5}{c}@{}}
    \toprule
    \multicolumn{2}{c}{\textbf{Visual alignment}} &
    \multicolumn{4}{c}{\textbf{Task score (\%)}} &
    \textbf{Mean} \\
    \cmidrule(lr){1-2}\cmidrule(lr){3-6}\cmidrule(l){7-7}
    \begin{tabular}{@{}c@{}}Image\\augmentation\end{tabular}
      & \begin{tabular}{@{}c@{}}Robot-arm\\rendering\end{tabular}
      & Close Laptop & Open Fridge & Pull Curtain & Close Drawer & \\
    \midrule
    $\times$ & $\times$ & 31.67 & 16.67 & 35.00 & \underline{50.00} & 33.33 \\
    $\times$ & $\checkmark$ & 51.67 & 23.33 & \underline{41.67} & --- & 38.89 \\
    $\checkmark$ & $\times$ & \textbf{61.67} & \underline{33.33} & 33.33 & \textbf{61.67} & \underline{47.50} \\
    \rowcolor{papertealtint}
    $\checkmark$ & $\checkmark$ & \underline{53.33} & \textbf{38.33} & \textbf{53.33} & --- & \textbf{48.33} \\
    \bottomrule
  \end{tabular}
\end{table}

Across the three arm tasks, rendering, augmentation, and their combination
score 38.89\%, 42.78\%, and 48.33\%, respectively, with unchanged action
labels. Their benefits are complementary but task-dependent: combining them
performs best on Open Fridge and Pull Curtain, while augmentation alone is
best on Close Laptop.

\section{Conclusion}
\label{sec:conclusion}

EgoHumanoid-V2 transfers coordinated whole-body loco-manipulation skills
from egocentric demonstrations through coarse-to-fine action alignment and
visual adaptation. With 200 human demonstrations per task, VLA policies
achieve a mean score of 51.67\% across four real-world tasks without
target-task robot demonstrations, demonstrating zero-shot skill transfer
on the tested tasks. The results support human motion as direct supervision
for coordinated humanoid skills.

\paragraph{Limitations and future work.}
\label{sec:discussion}
Alignment can still produce unstable motions, and IK and repeated rollouts
limit throughput. Stronger stability constraints and faster optimization
remain directions for improving reliability and processing efficiency.

\clearpage
\section*{AI Use Statement}
Generative AI tools assisted with literature discovery and synthesis, method
and experimental-design feedback, manuscript organization, drafting and editing,
reference formatting, figure development, and software development. The
authors reviewed this work and take responsibility for the final text, claims,
citations, code, and artifacts.

\section*{Reproducibility Statement}
Sections~\ref{sec:method} and~\ref{sec:experiments} describe the learning
interface and controlled comparisons. Appendices~\ref{sec:appendix}
and~\ref{sec:policy-evaluation} provide implementation and policy evaluation
details. Appendices~\ref{sec:action-ablation-protocol}
and~\ref{sec:online-ik-instability} report supplementary action-alignment
experiments and online-IK diagnostics.

\begingroup
\microtypesetup{activate=true}
\makeatletter
\long\def\BR@@lbibitem[#1]#2#3\par{%
  \let\backrefprint\BR@backrefprint
  \BRorg@bibitem[{#1}]{#2}\KeepBibliographyTail{#3}%
  \BR@backref{#2}}
\makeatother
\bibliographystyle{iclr2027_conference}
\bibliography{references}

\begin{thebibliography}{53}
\providecommand{\natexlab}[1]{#1}
\providecommand{\url}[1]{\texttt{#1}}
\expandafter\ifx\csname urlstyle\endcsname\relax
  \providecommand{\doi}[1]{doi: #1}\else
  \providecommand{\doi}{doi: \begingroup \urlstyle{rm}\Url}\fi

\bibitem[Allshire et~al.(2025)Allshire, Choi, Zhang, McAllister, Zhang, Kim, Darrell, Abbeel, Malik, and Kanazawa]{videomimic}
Arthur Allshire, Hongsuk Choi, Junyi Zhang, David McAllister, Anthony Zhang, Chung~Min Kim, Trevor Darrell, Pieter Abbeel, Jitendra Malik, and Angjoo Kanazawa.
\newblock Visual imitation enables contextual humanoid control.
\newblock In \emph{CoRL}, 2025.

\bibitem[Araujo et~al.(2026)Araujo, Ze, Xu, Wu, and Liu]{joao2025gmr}
Joao~Pedro Araujo, Yanjie Ze, Pei Xu, Jiajun Wu, and C.~Karen Liu.
\newblock {Retargeting Matters}: General motion retargeting for humanoid motion tracking.
\newblock In \emph{ICRA}, 2026.

\bibitem[Ben et~al.(2025)Ben, Jia, Zeng, Dong, Lin, and Pang]{homie}
Qingwei Ben, Feiyu Jia, Jia Zeng, Junting Dong, Dahua Lin, and Jiangmiao Pang.
\newblock {HOMIE}: Humanoid loco-manipulation with isomorphic exoskeleton cockpit.
\newblock In \emph{RSS}, 2025.

\bibitem[Bu et~al.(2025)Bu, Yang, Cai, Gao, Ren, Yao, Luo, and Li]{univla}
Qingwen Bu, Yanting Yang, Jisong Cai, Shenyuan Gao, Guanghui Ren, Maoqing Yao, Ping Luo, and Hongyang Li.
\newblock Learning to act anywhere with task-centric latent actions.
\newblock In \emph{RSS}, 2025.

\bibitem[Chen et~al.(2024)Chen, Hari, Dharmarajan, Xu, Vuong, and Goldberg]{chen2024mirage}
Lawrence~Yunliang Chen, Kush Hari, Karthik Dharmarajan, Chenfeng Xu, Quan Vuong, and Ken Goldberg.
\newblock Mirage: Cross-embodiment zero-shot policy transfer with cross-painting.
\newblock In \emph{RSS}, 2024.

\bibitem[Chen et~al.(2025)Chen, Xu, Dharmarajan, Cheng, Keutzer, Tomizuka, Vuong, and Goldberg]{chen2024rovi}
Lawrence~Yunliang Chen, Chenfeng Xu, Karthik Dharmarajan, Richard Cheng, Kurt Keutzer, Masayoshi Tomizuka, Quan Vuong, and Ken Goldberg.
\newblock {RoVi-Aug}: Robot and viewpoint augmentation for cross-embodiment robot learning.
\newblock In \emph{CoRL}, 2025.

\bibitem[Chen et~al.(2026)Chen, Ji, Cheng, Peng, Peng, and Wang]{gmt}
Zixuan Chen, Mazeyu Ji, Xuxin Cheng, Xuanbin Peng, Xue~Bin Peng, and Xiaolong Wang.
\newblock {GMT}: General motion tracking for humanoid whole-body control.
\newblock In \emph{IROS}, 2026.

\bibitem[Chi et~al.(2024)Chi, Xu, Pan, Cousineau, Burchfiel, Feng, Tedrake, and Song]{chi2024umi}
Cheng Chi, Zhenjia Xu, Chuer Pan, Eric Cousineau, Benjamin Burchfiel, Siyuan Feng, Russ Tedrake, and Shuran Song.
\newblock {Universal Manipulation Interface}: In-the-wild robot teaching without in-the-wild robots.
\newblock In \emph{RSS}, 2024.

\bibitem[Fu et~al.(2025)Fu, Zhao, Wu, Wetzstein, and Finn]{humanplus}
Zipeng Fu, Qingqing Zhao, Qi~Wu, Gordon Wetzstein, and Chelsea Finn.
\newblock {HumanPlus}: Humanoid shadowing and imitation from humans.
\newblock In \emph{CoRL}, 2025.

\bibitem[Grauman et~al.(2022)Grauman, Westbury, Byrne, Chavis, Furnari, Girdhar, Hamburger, Jiang, Liu, Liu, et~al.]{grauman2022ego4d}
Kristen Grauman, Andrew Westbury, Eugene Byrne, Zachary Chavis, Antonino Furnari, Rohit Girdhar, Jackson Hamburger, Hao Jiang, Miao Liu, Xingyu Liu, et~al.
\newblock {Ego4D}: Around the world in 3,000 hours of egocentric video.
\newblock In \emph{CVPR}, 2022.

\bibitem[Grauman et~al.(2024)Grauman, Westbury, Torresani, Kitani, Malik, Afouras, Ashutosh, Baiyya, Bansal, Boote, et~al.]{grauman2024egoexo4d}
Kristen Grauman, Andrew Westbury, Lorenzo Torresani, Kris Kitani, Jitendra Malik, Triantafyllos Afouras, Kumar Ashutosh, Vijay Baiyya, Siddhant Bansal, Bikram Boote, et~al.
\newblock {Ego-Exo4D}: Understanding skilled human activity from first- and third-person perspectives.
\newblock In \emph{CVPR}, 2024.

\bibitem[He et~al.(2024)He, Luo, Xiao, Zhang, Kitani, Liu, and Shi]{h2o}
Tairan He, Zhengyi Luo, Wenli Xiao, Chong Zhang, Kris Kitani, Changliu Liu, and Guanya Shi.
\newblock Learning human-to-humanoid real-time whole-body teleoperation.
\newblock In \emph{IROS}, 2024.

\bibitem[He et~al.(2025{\natexlab{a}})He, Gao, Xiao, Zhang, Wang, Wang, Luo, He, Sobanbabu, Pan, Yi, Qu, Kitani, Hodgins, Fan, Zhu, Liu, and Shi]{asap}
Tairan He, Jiawei Gao, Wenli Xiao, Yuanhang Zhang, Zi~Wang, Jiashun Wang, Zhengyi Luo, Guanqi He, Nikhil Sobanbabu, Chaoyi Pan, Zeji Yi, Guannan Qu, Kris Kitani, Jessica Hodgins, Linxi~"Jim" Fan, Yuke Zhu, Changliu Liu, and Guanya Shi.
\newblock {ASAP}: Aligning simulation and real-world physics for learning agile humanoid whole-body skills.
\newblock In \emph{RSS}, 2025{\natexlab{a}}.

\bibitem[He et~al.(2025{\natexlab{b}})He, Luo, He, Xiao, Zhang, Zhang, Kitani, Liu, and Shi]{omnih2o}
Tairan He, Zhengyi Luo, Xialin He, Wenli Xiao, Chong Zhang, Weinan Zhang, Kris~M Kitani, Changliu Liu, and Guanya Shi.
\newblock {OmniH2O}: Universal and dexterous human-to-humanoid whole-body teleoperation and learning.
\newblock In \emph{CoRL}, 2025{\natexlab{b}}.

\bibitem[Hoque et~al.(2026)Hoque, Huang, Yoon, Sivapurapu, and Zhang]{hoque2025egodex}
Ryan Hoque, Peide Huang, David~J Yoon, Mouli Sivapurapu, and Jian Zhang.
\newblock {EgoDex}: Learning dexterous manipulation from large-scale egocentric video.
\newblock In \emph{ICLR}, 2026.

\bibitem[Intelligence et~al.(2025)Intelligence, Black, Brown, Darpinian, Dhabalia, Driess, Esmail, Equi, Finn, Fusai, et~al.]{intelligence2025pi05}
Physical Intelligence, Kevin Black, Noah Brown, James Darpinian, Karan Dhabalia, Danny Driess, Adnan Esmail, Michael Equi, Chelsea Finn, Niccolo Fusai, et~al.
\newblock $\pi _{0.5}$: a vision-language-action model with open-world generalization.
\newblock In \emph{CoRL}, 2025.

\bibitem[Ji et~al.(2025)Ji, Peng, Liu, Li, Yang, Cheng, and Wang]{exbody2}
Mazeyu Ji, Xuanbin Peng, Fangchen Liu, Jialong Li, Ge~Yang, Xuxin Cheng, and Xiaolong Wang.
\newblock {ExBody2}: Advanced expressive humanoid whole-body control.
\newblock In \emph{RSS Workshop on Whole-Body Control and Bimanual Manipulation}, 2025.

\bibitem[Jiang et~al.(2026)Jiang, Chen, Bu, Chen, Shi, Zhang, Li, Suo, Wang, Peng, et~al.]{jiang2025wholebodyvla}
Haoran Jiang, Jin Chen, Qingwen Bu, Li~Chen, Modi Shi, Yanjie Zhang, Delong Li, Chuanzhe Suo, Chuang Wang, Zhihui Peng, et~al.
\newblock {WholeBodyVLA}: Towards unified latent vla for whole-body loco-manipulation control.
\newblock In \emph{ICLR}, 2026.

\bibitem[Kareer et~al.(2025)Kareer, Patel, Punamiya, Mathur, Cheng, Wang, Hoffman, and Xu]{kareer2025egomimic}
Simar Kareer, Dhruv Patel, Ryan Punamiya, Pranay Mathur, Shuo Cheng, Chen Wang, Judy Hoffman, and Danfei Xu.
\newblock {EgoMimic}: Scaling imitation learning via egocentric video.
\newblock In \emph{ICRA}, 2025.

\bibitem[Kareer et~al.(2026)Kareer, Pertsch, Darpinian, Hoffman, Xu, Levine, Finn, and Nair]{kareer2025emergence}
Simar Kareer, Karl Pertsch, James Darpinian, Judy Hoffman, Danfei Xu, Sergey Levine, Chelsea Finn, and Suraj Nair.
\newblock Emergence of human to robot transfer in vision-language-action models.
\newblock In \emph{RSS}, 2026.

\bibitem[Li et~al.(2025{\natexlab{a}})Li, Cheng, Huang, Yang, Qiu, and Wang]{amo}
Jialong Li, Xuxin Cheng, Tianshu Huang, Shiqi Yang, Ri-Zhao Qiu, and Xiaolong Wang.
\newblock {AMO}: Adaptive motion optimization for hyper-dexterous humanoid whole-body control.
\newblock In \emph{RSS}, 2025{\natexlab{a}}.

\bibitem[Li et~al.(2025{\natexlab{b}})Li, Xue, Ren, and Bo]{li2025diffueraser}
Xiaowen Li, Haolan Xue, Peiran Ren, and Liefeng Bo.
\newblock {DiffuEraser}: A diffusion model for video inpainting.
\newblock \emph{arXiv preprint arXiv:2501.10018}, 2025{\natexlab{b}}.

\bibitem[Lin et~al.(2026)Lin, Arora, Mercat, Nishimura, Shah, Xu, Zhang, Zolotas, Angeles, Pfannenstiehl, Beaulieu, and Barreiros]{lin2026cotraininglbm}
Fanqi Lin, Kushal Arora, Jean Mercat, Haruki Nishimura, Paarth Shah, Chen Xu, Mengchao Zhang, Mark Zolotas, Maya Angeles, Owen Pfannenstiehl, Andrew Beaulieu, and Jose Barreiros.
\newblock A systematic study of data modalities and strategies for co-training large behavior models for robot manipulation.
\newblock In \emph{RSS}, 2026.

\bibitem[Liu et~al.(2026)Liu, Li, Ma, Wu, Tan, Ouyang, Su, and Zhu]{liu2026rdt2}
Songming Liu, Bangguo Li, Kai Ma, Lingxuan Wu, Hengkai Tan, Xiao Ouyang, Hang Su, and Jun Zhu.
\newblock {RDT2}: Exploring the scaling limit of {UMI} data towards zero-shot cross-embodiment generalization.
\newblock In \emph{ICML}, 2026.

\bibitem[Luo et~al.(2026)Luo, Yuan, Wang, Li, Chen, Casta{\~n}eda, Cao, Li, Minor, Ben, et~al.]{luo2025sonic}
Zhengyi Luo, Ye~Yuan, Tingwu Wang, Chenran Li, Sirui Chen, Fernando Casta{\~n}eda, Zi-Ang Cao, Jiefeng Li, David Minor, Qingwei Ben, et~al.
\newblock {SONIC}: Supersizing motion tracking for natural humanoid whole-body control.
\newblock \emph{Science Robotics}, 2026.
\newblock In press.

\bibitem[Ma et~al.(2023)Ma, Sodhani, Jayaraman, Bastani, Kumar, and Zhang]{ma2022vip}
Yecheng~Jason Ma, Shagun Sodhani, Dinesh Jayaraman, Osbert Bastani, Vikash Kumar, and Amy Zhang.
\newblock {VIP}: Towards universal visual reward and representation via value-implicit pre-training.
\newblock In \emph{ICLR}, 2023.

\bibitem[Nai et~al.(2026)Nai, Zheng, Zhao, Zhu, Dai, Chen, Hu, Hu, Zhang, Wen, and Gao]{nai2026humi}
Ruiqian Nai, Boyuan Zheng, Junming Zhao, Haodong Zhu, Sicong Dai, Zunhao Chen, Yihang Hu, Yingdong Hu, Tong Zhang, Chuan Wen, and Yang Gao.
\newblock Humanoid manipulation interface: Humanoid whole-body manipulation from robot-free demonstrations.
\newblock In \emph{CoRL}, 2026.

\bibitem[Nair et~al.(2023)Nair, Rajeswaran, Kumar, Finn, and Gupta]{nair2022r3m}
Suraj Nair, Aravind Rajeswaran, Vikash Kumar, Chelsea Finn, and Abhinav Gupta.
\newblock {R3M}: A universal visual representation for robot manipulation.
\newblock In \emph{CoRL}, 2023.

\bibitem[Pan et~al.(2026{\natexlab{a}})Pan, Wang, Qi, Liu, Bharadhwaj, Sharma, Wu, Shi, Malik, and Hogan]{pan2025spider}
Chaoyi Pan, Changhao Wang, Haozhi Qi, Zixi Liu, Homanga Bharadhwaj, Akash Sharma, Tingfan Wu, Guanya Shi, Jitendra Malik, and Francois Hogan.
\newblock {SPIDER}: Scalable physics-informed dexterous retargeting.
\newblock In \emph{IROS}, 2026{\natexlab{a}}.

\bibitem[Pan et~al.(2026{\natexlab{b}})Pan, Qiao, Chen, Chitta, Pan, Mai, Bu, Zheng, Zhao, Luo, and Li]{pan2025ams}
Yixuan Pan, Ruoyi Qiao, Li~Chen, Kashyap Chitta, Liang Pan, Haoguang Mai, Qingwen Bu, Cunyuan Zheng, Hao Zhao, Ping Luo, and Hongyang Li.
\newblock {Agility Meets Stability}: Versatile humanoid control with heterogeneous data.
\newblock In \emph{ICRA}, 2026{\natexlab{b}}.

\bibitem[Punamiya et~al.(2025)Punamiya, Patel, Aphiwetsa, Kuppili, Zhu, Kareer, Hoffman, and Xu]{punamiya2025egobridge}
Ryan Punamiya, Dhruv Patel, Patcharapong Aphiwetsa, Pranav Kuppili, Lawrence~Y. Zhu, Simar Kareer, Judy Hoffman, and Danfei Xu.
\newblock {EgoBridge}: Domain adaptation for generalizable imitation from egocentric human data.
\newblock In \emph{NeurIPS}, 2025.

\bibitem[Qi et~al.(2026)Qi, Chen, Wang, Lin, Lian, Zhang, Yu, Wang, and Yi]{qi2026humanoidgpt}
Zekun Qi, Xuchuan Chen, Jilong Wang, Chenghuai Lin, Yunrui Lian, Wenyao Zhang, Xinqiang Yu, He~Wang, and Li~Yi.
\newblock Humanoid generative pre-training for zero-shot motion tracking.
\newblock In \emph{CVPR}, 2026.

\bibitem[Qiu et~al.(2025)Qiu, Yang, Cheng, Chawla, Li, He, Yan, Yoon, Hoque, Paulsen, et~al.]{qiu2025hat}
Ri-Zhao Qiu, Shiqi Yang, Xuxin Cheng, Chaitanya Chawla, Jialong Li, Tairan He, Ge~Yan, David~J Yoon, Ryan Hoque, Lars Paulsen, et~al.
\newblock Humanoid policy {\textasciitilde} human policy.
\newblock In \emph{CoRL}, 2025.

\bibitem[Ravi et~al.(2025)Ravi, Gabeur, Hu, Hu, Ryali, Ma, Khedr, R{\"a}dle, Rolland, Gustafson, Mintun, Pan, Alwala, Carion, Wu, Girshick, Doll{\'a}r, and Feichtenhofer]{ravi2024sam2}
Nikhila Ravi, Valentin Gabeur, Yuan-Ting Hu, Ronghang Hu, Chaitanya Ryali, Tengyu Ma, Haitham Khedr, Roman R{\"a}dle, Chloe Rolland, Laura Gustafson, Eric Mintun, Junting Pan, Kalyan~Vasudev Alwala, Nicolas Carion, Chao-Yuan Wu, Ross Girshick, Piotr Doll{\'a}r, and Christoph Feichtenhofer.
\newblock {SAM 2}: Segment anything in images and videos.
\newblock In \emph{ICLR}, 2025.

\bibitem[Shi et~al.(2026)Shi, Peng, Chen, Jiang, Li, Luo, Huang, Li, and Chen]{shi2026egohumanoid}
Modi Shi, Shijia Peng, Jin Chen, Haoran Jiang, Tianyu Li, Ping Luo, Di~Huang, Hongyang Li, and Li~Chen.
\newblock Unlocking in-the-wild loco-manipulation with robot-free egocentric demonstration.
\newblock In \emph{RSS}, 2026.

\bibitem[Tao et~al.(2025)Tao, Srirama, Liu, Shaw, and Pathak]{tao2025dexwild}
Tony Tao, Mohan~Kumar Srirama, Jason~Jingzhou Liu, Kenneth Shaw, and Deepak Pathak.
\newblock {DexWild}: Dexterous human interactions for in-the-wild robot policies.
\newblock In \emph{RSS}, 2025.

\bibitem[Wang et~al.(2026)Wang, Yu, Hu, Zhang, Li, and Luo]{wang2026bifrostumi}
Hongwu Wang, Chenhao Yu, Youhao Hu, Jiachen Zhang, Yuanyuan Li, and Shaqi Luo.
\newblock {BifrostUMI}: Bridging robot-free demonstrations and humanoid whole-body manipulation.
\newblock \emph{arXiv preprint arXiv:2605.03452}, 2026.

\bibitem[Wei et~al.(2026)Wei, Jing, Li, Zhao, Mao, Ni, He, Zang, Liu, Kang, Liu, Yuan, Pavone, Huang, and Wang]{wei2026psi0}
Songlin Wei, Hongyi Jing, Boqian Li, Zhenyu Zhao, Jiageng Mao, Zhenhao Ni, Sicheng He, Sheng Zang, Xiawei Liu, Kaidi Kang, Jie Liu, Weiduo Yuan, Marco Pavone, Di~Huang, and Yue Wang.
\newblock {$\Psi_0$}: An open foundation model towards universal humanoid loco-manipulation.
\newblock In \emph{RSS}, 2026.

\bibitem[Wu et~al.(2024)Wu, Jing, Cheang, Chen, Xu, Li, Liu, Li, and Kong]{wu2023gr1}
Hongtao Wu, Ya~Jing, Chilam Cheang, Guangzeng Chen, Jiafeng Xu, Xinghang Li, Minghuan Liu, Hang Li, and Tao Kong.
\newblock Unleashing large-scale video generative pre-training for visual robot manipulation.
\newblock In \emph{ICLR}, 2024.

\bibitem[Xu et~al.(2025)Xu, Zhang, Hou, Xu, Fan, Veloso, and Song]{xu2025dexumi}
Mengda Xu, Han Zhang, Yifan Hou, Zhenjia Xu, Linxi Fan, Manuela Veloso, and Shuran Song.
\newblock {DexUMI}: Using human hand as the universal manipulation interface for dexterous manipulation.
\newblock In \emph{CoRL}, 2025.

\bibitem[Xu et~al.(2026{\natexlab{a}})Xu, Yin, Zhou, Zhang, Chen, and Hu]{xu2026glori}
Qingyao Xu, Sheng Yin, Zibo Zhou, Ya~Zhang, Siheng Chen, and Yue Hu.
\newblock {GLoRI}: Closed-loop whole-body tracking with global-local reference interaction for humanoid loco-manipulation.
\newblock \emph{arXiv preprint arXiv:2609.05994}, 2026{\natexlab{a}}.

\bibitem[Xu et~al.(2026{\natexlab{b}})Xu, Park, Zhang, Cousineau, Bhat, Barreiros, Wang, Bohg, and Song]{xu2026hommi}
Xiaomeng Xu, Jisang Park, Han Zhang, Eric Cousineau, Aditya Bhat, Jose Barreiros, Dian Wang, Jeannette Bohg, and Shuran Song.
\newblock {HoMMI}: Learning whole-body mobile manipulation from human demonstrations.
\newblock In \emph{RSS}, 2026{\natexlab{b}}.

\bibitem[Yang et~al.(2026{\natexlab{a}})Yang, Bao, Xin, Song, Tian, Zhao, Wang, and Li]{yang2026zerowbc}
Haoran Yang, Jiacheng Bao, Yucheng Xin, Haoming Song, Yuyang Tian, Bin Zhao, Dong Wang, and Xuelong Li.
\newblock {ZeroWBC}: Learning natural whole-body humanoid interaction from human egocentric data.
\newblock \emph{arXiv preprint arXiv:2603.09170}, 2026{\natexlab{a}}.

\bibitem[Yang et~al.(2026{\natexlab{b}})Yang, Huang, Wu, Kanazawa, Abbeel, Sferrazza, Liu, Duan, and Shi]{yang2025omniretarget}
Lujie Yang, Xiaoyu Huang, Zhen Wu, Angjoo Kanazawa, Pieter Abbeel, Carmelo Sferrazza, C~Karen Liu, Rocky Duan, and Guanya Shi.
\newblock {OmniRetarget}: Interaction-preserving data generation for humanoid whole-body loco-manipulation and scene interaction.
\newblock In \emph{ICRA}, 2026{\natexlab{b}}.

\bibitem[Ye et~al.(2025)Ye, Jang, Jeon, Joo, Yang, Peng, Mandlekar, Tan, Chao, Lin, et~al.]{ye2024lapa}
Seonghyeon Ye, Joel Jang, Byeongguk Jeon, Sejune Joo, Jianwei Yang, Baolin Peng, Ajay Mandlekar, Reuben Tan, Yu-Wei Chao, Bill~Yuchen Lin, et~al.
\newblock Latent action pretraining from videos.
\newblock In \emph{ICLR}, 2025.

\bibitem[Ye et~al.(2026)Ye, Ge, Zheng, Gao, Yu, Kurian, Indupuru, Tan, Zhu, Xiang, Malik, Lee, Liang, Ranawaka, Gu, Xu, Wang, Hu, Narayan, Bjorck, Wang, Kim, Niu, Zheng, Xie, Wu, Wang, Julian, Xu, Du, Chebotar, Reed, Kautz, Zhu, Fan, and Jang]{ye2026dreamzero}
Seonghyeon Ye, Yunhao Ge, Kaiyuan Zheng, Shenyuan Gao, Sihyun Yu, George Kurian, Suneel Indupuru, You~Liang Tan, Chuning Zhu, Jiannan Xiang, Ayaan Malik, Kyungmin Lee, William Liang, Nadun Ranawaka, Jiasheng Gu, Yinzhen Xu, Guanzhi Wang, Fengyuan Hu, Avnish Narayan, Johan Bjorck, Jing Wang, Gwanghyun Kim, Dantong Niu, Ruijie Zheng, Yuqi Xie, Jimmy Wu, Qi~Wang, Ryan Julian, Danfei Xu, Yilun Du, Yevgen Chebotar, Scott Reed, Jan Kautz, Yuke Zhu, Linxi Fan, and Joel Jang.
\newblock World action models are zero-shot policies.
\newblock In \emph{CoRL}, 2026.

\bibitem[Yuan et~al.(2026)Yuan, Zhou, Liu, Hu, Wang, Yi, Wen, Zhang, and Gao]{yuan2025motiontrans}
Chengbo Yuan, Rui Zhou, Mengzhen Liu, Yingdong Hu, Shengjie Wang, Li~Yi, Chuan Wen, Shanghang Zhang, and Yang Gao.
\newblock {MotionTrans}: Human vr data enable motion-level learning for robotic manipulation policies.
\newblock In \emph{ICRA}, 2026.

\bibitem[Ze et~al.(2025)Ze, Chen, Araújo, ang Cao, Peng, Wu, and Liu]{twist}
Yanjie Ze, Zixuan Chen, João~Pedro Araújo, Zi~ang Cao, Xue~Bin Peng, Jiajun Wu, and C.~Karen Liu.
\newblock {TWIST}: Teleoperated whole-body imitation system.
\newblock In \emph{CoRL}, 2025.

\bibitem[Zeng et~al.(2024)Zeng, Bu, Wang, Xia, Chen, Dong, Song, Wang, Hu, Luo, et~al.]{mpi}
Jia Zeng, Qingwen Bu, Bangjun Wang, Wenke Xia, Li~Chen, Hao Dong, Haoming Song, Dong Wang, Di~Hu, Ping Luo, et~al.
\newblock Learning manipulation by predicting interaction.
\newblock In \emph{RSS}, 2024.

\bibitem[Zhao et~al.(2026)Zhao, Ze, Wang, Liu, Abbeel, Shi, and Duan]{zhao2025resmimic}
Siheng Zhao, Yanjie Ze, Yue Wang, C~Karen Liu, Pieter Abbeel, Guanya Shi, and Rocky Duan.
\newblock {ResMimic}: From general motion tracking to humanoid whole-body loco-manipulation via residual learning.
\newblock In \emph{ICRA}, 2026.

\bibitem[Zhaxizhuoma et~al.(2025)Zhaxizhuoma, Liu, Guan, Jia, Wu, Liu, Wang, Liang, CHEN, Zhang, et~al.]{zhaxizhuoma2025fastumi}
Zhaxizhuom Zhaxizhuoma, Kehui Liu, Chuyue Guan, Zhongjie Jia, Ziniu Wu, Xin Liu, Tianyu Wang, Shuai Liang, Pengan CHEN, Pingrui Zhang, et~al.
\newblock {FastUMI}: A scalable and hardware-independent universal manipulation interface with dataset.
\newblock In \emph{CoRL}, 2025.

\bibitem[Zheng et~al.(2026)Zheng, Niu, Xie, Wang, Xu, Jiang, Castaneda, Hu, Tan, Fu, Darrell, Huang, Zhu, Xu, and Fan]{zheng2026egoscale}
Ruijie Zheng, Dantong Niu, Yuqi Xie, Jing Wang, Mengda Xu, Yunfan Jiang, Fernando Castaneda, Fengyuan Hu, You~Liang Tan, Letian Fu, Trevor Darrell, Furong Huang, Yuke Zhu, Danfei Xu, and Linxi Fan.
\newblock {EgoScale}: Scaling dexterous manipulation with diverse egocentric human data.
\newblock In \emph{CoRL}, 2026.

\bibitem[Zhong et~al.(2026)Zhong, Sun, Wen, Li, Cheng, Dai, Zeng, Lu, Zhu, and Xu]{zhong2025humanoidexo}
Rui Zhong, Yizhe Sun, Junjie Wen, Jinming Li, Chuang Cheng, Wei Dai, Zhiwen Zeng, Huimin Lu, Yichen Zhu, and Yi~Xu.
\newblock {HumanoidExo}: Scalable whole-body humanoid manipulation via wearable exoskeleton.
\newblock In \emph{ICRA}, 2026.

\end{thebibliography}
\endgroup

\clearpage
\appendix
\section*{\normalfont\Large\bfseries\itshape Appendix}

\section{Methodology}
\label{sec:appendix}

\subsection{Data Preparation and Validation}
\label{sec:data-validation}
GoPro RGB, PICO whole-body tracking, and hand-pose signals are synchronized
into 50~Hz image--trajectory pairs. Signal validation and heading calibration
precede motion initialization $H$; the resulting reference then passes through
$K$ and $D$. Accepted replays supply the states and actions paired with aligned
images for VLA post-training. Each sample uses the 43-dimensional state and
76-dimensional action interface in Eq.~\ref{eq:runtime-interface}.

Validation follows the same processing order. Checks of timestamp consistency
and agreement between video and trajectory
frame counts establish whether the source streams can be paired.
Endpoint tracking error, changes in joint angles between frames, base tilt,
and pelvis displacement assess
the resulting robot execution. Without an external pulse, PICO--GoPro timing
consistency measures repeatability rather than absolute synchronization error.

\subsection{Action Alignment Details}
\label{sec:alignment-implementation}
Both stages use fixed targets from the human recording: kinematic alignment
adjusts the robot reference, and dynamics-aware alignment refines its motion tokens.

\paragraph{Target construction and registration.}
We first remove world translation and heading by expressing each world-space
point $p_t^W$ in a pelvis-centered frame:
\begin{equation}
p_t^L=R_z(\psi_t)^\top(p_t^W-p_{\mathrm{pelvis},t}^W),
\label{eq:local-frame}
\end{equation}
where $\psi_t$ is pelvis yaw. The frame keeps its $z$ axis vertical, so it does
not rotate with body tilt. Human palm targets retain their local horizontal
coordinates and orientation. To register height, we set
$y_{t,z}=p^{\mathrm H,L}_{t,z}+h_t^{\mathrm H}-h_t^{\mathrm R,0}$, where
$h_t^{\mathrm H}$ and $h_t^{\mathrm R,0}$ are pelvis heights above the floor
in the human recording and initial robot rollout. This registration is
fixed across refinement rounds. Human-conditioned controller execution
supplies the initial joint reference, and independent token replay measures
tracking error against the registered targets.

Foot targets approximate each sole center by the midpoint between the
human ankle and foot keypoints. We estimate the ground height as the median,
over the demonstration, of the lowest ankle or foot keypoint in each frame.
We similarly estimate the ground height from the lower of the two sole-center
proxies. We subtract the proxy-based estimate minus the keypoint-based
estimate from the vertical coordinate of each proxy. This vertical correction
is constant throughout the demonstration
and preserves changes in foot height. We then apply the same pelvis-height
registration used for the hands. The resulting targets approximate sole
centers; they are not measured contact points. Both $K$ and $D$ target bilateral hand poses by
default; foot-only tasks such as Close Drawer instead target foot positions
and omit the orientation objective.

\paragraph{Kinematic reference correction ($K$).}
The kinematic stage adjusts the initial robot trajectory toward these targets
before encoding it as motion tokens. For hand tasks, the optimization acts
only on the arms and waist. Both IK solvers bound deviations from the initial
reference, retaining the original motion as the anchor for correction.
The corrected reference is then encoded and replayed to obtain the execution
residual addressed by $D$.

\paragraph{Dynamics-aware refinement ($D$).}
The dynamics-aware stage uses the response prediction in
Eq.~\ref{eq:response-prediction} to compensate for the measured execution
residual. Within each local solve, the context $c^k$ and observation history
$s^k$ remain fixed, and the task metric $Q_\tau$ weights endpoint errors:
\begin{equation}
z^{k+1}=\arg\min_{z\in\mathcal{N}(z^k)}
\|\widehat{\Delta e}(z)-r^k\|_{Q_\tau}^2
+\mathcal{L}_{\mathrm{reg}}(z,z^k).
\end{equation}
The neighborhood $\mathcal{N}(z^k)$ limits how far the tokens can change
from their current values $z^k$. The regularizer
combines a joint-command penalty $\|\Delta u(z)\|_{W_\tau}^2$, token-change
and temporal penalties, and an IK correction prior. Joint-preservation
weights for the legs, waist, and arms are $(20,20,0.1)$ for hand tasks and
$(0.1,20,20)$ for foot-only tasks. These are soft penalties on proposed
corrections; they do not freeze body parts during execution.

\paragraph{Optimization schedule and output.}
Each refinement round locally linearizes endpoint geometry and performs
400 Adam steps through the frozen decoder and MLP response model. A fresh
rollout evaluates the candidate and supplies the next round's execution
context. The default schedule comprises one $K$ round, seven regular $D$
rounds, and one conservative $D$ round. The recorded candidates then enter
the selection and independent validation procedure in
Appendix~\ref{sec:data-acceptance}.

\subsection{Response Model and Training}
\label{sec:response-training}
The response model predicts how a kinematic command change affects execution.

\paragraph{Paired-rollout supervision.}
Response supervision uses G1 motion trajectories from
BONES-SEED\footnote{\url{https://huggingface.co/datasets/bones-studio/seed}}.
Reference and perturbed tokens are replayed at 50~Hz from identical
simulator states and controller histories. Perturbations combine smooth
random changes and token-refinement directions. The input
is the Jacobian-based endpoint change between commands decoded under the
same baseline observation; the target is the difference between the endpoint
poses executed in the two
rollouts one policy step later. Training uses 250 motions from 68 capture dates (345,080 frame--perturbation
pairs); validation uses 13 motions from seven other dates (20,420 pairs).

\paragraph{Inputs and temporal context.}
Position and rotation changes for both palms and soles form a 24-D
intervention input. The current input and five exponentially smoothed versions of that input,
with time constants of 0.02, 0.1, 0.5, 2, and 5~s, form a 144-D causal
history. The predictor also conditions on 93 robot-state features and the
64-D reference token. These fixed baseline features constitute $c^k$;
they and the causal history are omitted from the compact notation in
Eq.~\ref{eq:response-prediction}.

\paragraph{Architecture and training objective.}
A frozen linear response and an MLP residual jointly predict the
24-dimensional endpoint response. The MLP has two hidden layers of width
64 and SiLU activations. Subtracting its output at zero history enforces
zero response when the intervention history is zero. Inputs and targets
are normalized using training-set statistics, and the objective combines
normalized mean squared error with a squared residual penalty of weight 1.
Random perturbations receive 75\% of the training weight; the two
refinement-derived direction families share the remaining 25\%. Within each family, weights balance capture dates and source motions.

\paragraph{Optimization and use during refinement.}
We train with AdamW at learning rate $3\times10^{-5}$, weight decay $10^{-4}$,
batch size 512, and gradient clipping at norm 5. Training lasts at most
150 epochs, and the checkpoint with the lowest validation loss is retained.
During token optimization, the model weights and baseline features stay
fixed; gradients traverse the candidate-dependent history rather than the
simulator. Palm tasks blend the learned and unit responses with a
learned-response weight of 0.15. Foot-only tasks use the learned position
response without an orientation objective. Development rollouts set these task-specific configurations, which remain fixed.

\subsection{Motion Selection and Validation}
\label{sec:data-acceptance}
Validation checks accuracy, execution consistency, and stability. Limits are
5~cm position error, $15^\circ$ orientation error where applicable,
0.10~m pelvis-height drop, and $25^\circ$ base tilt.

Candidates recorded during initialization, kinematic correction, and
refinement that satisfy
these checks are ranked first by their largest normalized endpoint error,
then by their mean normalized error. They are independently
replayed in that order. The selected candidate may be the initial rollout $H$ or the corrected
reference replay $H+K$ if later candidates rank lower or fail validation; an
episode is rejected if no candidate passes.
The accepted replay supplies the state--action pairs.

Task-specific settings also control preprocessing of refrigerator
trajectories, hand commands for pulling curtains, and the handling of
curtain sequence endings. All tasks use the
same low-level controller checkpoint without target-task training or
fine-tuning.

\section{Policy Training and Evaluation}
\label{sec:policy-evaluation}
The policy experiments test whether the aligned supervision supports
physical task execution.

\subsection{Post-Training Configuration}
\label{sec:post-training-config}
All controlled comparisons post-train $\pi_{0.5}$ from \texttt{pi05\_base}
in bfloat16 for 30,000 steps with global batch size 256. We use AdamW,
gradient clipping at 1.0, and weight decay $10^{-10}$. The learning rate
warms up for 1,000 steps to $2.5\times10^{-5}$, then follows a cosine decay
to $2.5\times10^{-6}$.

\paragraph{Visual preprocessing.}
\label{sec:visual-training-details}
Robot-arm rendering is performed offline as described in
Section~\ref{sec:visual-alignment}. During post-training, random resized
crops, in-plane rotation, and photometric jitter augment the images.
Deployment applies standard preprocessing to robot images only.

\subsection{Training Data and Controlled Comparisons}
\label{sec:controlled-comparisons}
The default human-data condition combines $H+K+D_s$ action labels,
robot-arm rendering, and image augmentation. Foot-operated Close Drawer
omits arm rendering. Across conditions, the policy backbone, action
interface, and evaluation protocol are held fixed; the following comparisons
vary the data budget or supervision component.

\paragraph{Data-budget comparisons.}
Q1 uses 25, 50, 100, or 200 human demonstrations per task. Q2 trains separate
human-data and robot-data policies at the budgets in
Section~\ref{sec:data-efficiency}. Action and visual ablations each use
200 demonstrations per task.

\paragraph{Supervision ablations.}
In Table~\ref{tab:action-transfer}, the $H$ and $H+K$ conditions use their
respective stage-specific action labels, while the full condition uses
validated historical candidates (Appendix~\ref{sec:data-acceptance}).
Visual preprocessing and optimization settings are identical across these
action-label comparisons. Visual ablations instead vary the visual
preprocessing components under the shared training configuration.

\subsection{Task Scoring Criteria}
\label{sec:task-success}
Each task consists of three sequential binary-scored subtasks.
Evaluation stops at the first failure, and all remaining subtasks receive
zero. A trial therefore scores $0$, $1/3$, $2/3$, or $1$. Each condition
includes 20 attempted trials, and the task score is
\begin{equation}
\mathrm{Task\ score}\ (\%) = \frac{100}{20}
\sum_{i=1}^{20}\left(\frac{1}{3}\sum_{j=1}^{3}s_{ij}\right),
\qquad s_{ij}\in\{0,1\}.
\label{eq:task-score}
\end{equation}
Table~\ref{tab:task-scoring-stages} lists the ordered criteria for each task.
The stages distinguish reaching or establishing contact, performing the
interaction, and completing it.

\begin{table}[!htbp]
\centering
\caption{Ordered binary scoring criteria for physical policy evaluation.}
\label{tab:task-scoring-stages}
\papertablestyle
\begin{tabular}{@{}p{0.13\linewidth}p{0.27\linewidth}p{0.27\linewidth}p{0.24\linewidth}@{}}
\toprule
Task & Stage 1 & Stage 2 & Stage 3 \\
\midrule
Close Laptop & Right hand reaches a suitable position above the lid.
& Right hand makes an effective downward press on the lid. & Lid is fully closed. \\
Open Fridge & Left hand approaches the door gap.
& Left hand hooks into the gap. & Refrigerator door opens. \\
Pull Curtain & Right hand grasps the curtain.
& Hand pulls the curtain. & Curtain is pulled open by one meter. \\
Close Drawer & Right foot lifts, with the toe in front of the lowest drawer's outer face.
& Right foot pushes the drawer inward. & Drawer is fully closed. \\
\bottomrule
\end{tabular}
\end{table}

\subsection{Result Aggregation}
\label{sec:result-reporting}
Task scores summarize partial completion rather than binary full-task
success. Cross-task means give equal weight to the available task scores;
visual-ablation rows with arm rendering exclude Close Drawer. No error bars
are reported.

The four policies are task-specific. Starting-pose and object variations
therefore assess within-task execution, rather than unseen-task
generalization by a single policy.

\section{Supplementary Action-Alignment Experiments}
\label{sec:action-ablation-protocol}
The supplementary experiments evaluate how alignment stages, motion
initialization, the number of refinement rounds, and the learned response
model affect tracking accuracy, motion quality, and processing time on the
four evaluation tasks. A separate 50-task study examines whether these
trends extend to a broader set of tasks.

\subsection{Shared Protocol and Metrics}
\label{sec:offline-metrics}
The four-task studies use the same fixed random sample of 25 training
trajectories per task (Section~\ref{sec:experimental-setup}). Targets,
registration, controller checkpoint, reset states, and evaluation windows
are matched, so the comparisons isolate changes in the alignment procedure.

\paragraph{Endpoint metrics and aggregation.}
Position error is measured in centimeters and orientation error in degrees.
For each trajectory, errors are averaged over frames and relevant endpoints;
trajectory means are then averaged within each task, and cross-task means
weight tasks equally. Position includes both palms for the three hand tasks
and both soles for Close Drawer. Orientation is evaluated only for the three
hand tasks. Missing measurements receive no numerical error and are not
replaced by a more favorable round. We average and rank unrounded values, then report two decimal places.

\paragraph{Candidate selection across comparisons.}
The stage comparison selects from all recorded candidates, including
initialization, kinematic correction, and refinement; the initialization
comparison selects only from post-$K$ refinement rounds; and the depth
curves report the candidate produced at each specified round, without
selecting an earlier candidate.

\subsection{Stage Comparison and Historical Selection}
\label{sec:stage-selection}
The offline error block in Table~\ref{tab:action-transfer} asks how much
accuracy is available from the recorded optimization history. Its candidate
set includes $H$, $H+K$, and all recorded $D$ rounds. For the three hand
tasks, we select a candidate by minimizing
\begin{equation}
S=0.75\,\frac{e_{\mathrm{pos}}}{5\,\mathrm{cm}}
 +0.25\,\frac{e_{\mathrm{ori}}}{15^\circ}.
\label{eq:stage-selection}
\end{equation}
Here, $e_{\mathrm{pos}}$ and $e_{\mathrm{ori}}$ are the mean bilateral palm
errors of the same candidate. Close Drawer instead minimizes mean bilateral
sole-position error. Because the initial and kinematically corrected
references remain eligible, the selected $H+K+D_s$ output may retain $H$
or $H+K$.

Recorded measurements are used without validation replay or acceptance filtering.

\subsection{Motion Initialization and Whole-Body Motion Quality}
\label{sec:initialization-details}
This comparison tests whether the initial reference affects accuracy and
whole-body motion quality after a shared refinement budget. We compare
SONIC Rollout, GMR retargeting, and HuMI-style whole-body
IK~\citep{nai2026humi}, adapted to our 29-joint model. Every reference is
encoded and replayed through the same SONIC checkpoint, then receives one
$K$ round and eight $D$ rounds. The comparison isolates initialization;
it does not reproduce the cited methods' complete pipelines.

\paragraph{Candidate selection.}
For each initializer, $D_s$ is selected only from $D^1$--$D^8$ after $K$.
For hand tasks, position errors are divided by 5\,cm and orientation errors
by 15$^\circ$. Candidates are ranked by the largest of these normalized
errors across both palms; ties are resolved by mean normalized error and
then iteration index. Close Drawer uses the analogous bilateral
sole-position rule without orientation. No training-data acceptance filter
is applied. Table~\ref{tab:initialization} reports task-averaged results,
while Tables~\ref{tab:initialization-taskwise}
and~\ref{tab:motion-quality-taskwise} give the corresponding task-wise errors
and motion diagnostics.

\paragraph{Motion-quality metrics.}
\label{sec:motion-quality-metrics}
Tables~\ref{tab:initialization} and~\ref{tab:motion-quality-taskwise}
report four complementary diagnostics.
\textbf{Pelvis drop} is the maximum decrease in pelvis height from its
initial value, in cm. \textbf{Tilt} is the maximum angle between the base's
vertical axis and world vertical. \textbf{Joint step} is the largest absolute
change in any of the 29 body-joint angles between consecutive replay frames,
measuring abrupt joint motion rather than walking step length.
\textbf{Extra yaw} compares robot and human heading changes. We unwrap each
yaw trajectory and subtract its initial angle, yielding $\psi^r$ and $\psi^h$.
The metric is
\begin{equation}
e_{\mathrm{yaw}}=\max\left\{
\left|\min_t\psi_t^r-\min_t\psi_t^h\right|,\;
\left|\max_t\psi_t^r-\max_t\psi_t^h\right|,\;
\left|\psi_T^r-\psi_T^h\right|\right\}.
\label{eq:extra-yaw}
\end{equation}
This measures differences in minimum, maximum, and final heading changes.
Angles are in degrees. Trajectory metrics are averaged within each task,
then equally across tasks.

\begin{table}[t]
  \centering
  \caption{\textbf{Task-wise endpoint error for motion initialization.}
  Position/orientation errors at $D_s$ are in cm/$^\circ$, averaged over 25
  episodes per task. Drawer orientation is not evaluated. Lower is better;
  bold and underline denote the best and second-best values for each metric.}
  \label{tab:initialization-taskwise}
  \papertablestyle
  \begin{tabular}{@{}lccc@{}}
    \toprule
    Task & SONIC Rollout & Retargeting & Whole-Body IK \\
    \midrule
    Close Laptop & 3.04 / 9.50 & \textbf{2.67} / \underline{6.33}
      & \underline{2.75} / \textbf{6.19} \\
    Open Fridge & \textbf{2.76} / \underline{5.77}
      & \underline{3.15} / \textbf{5.74} & 3.68 / 7.25 \\
    Pull Curtain & \textbf{2.02} / \underline{4.85}
      & \underline{2.05} / \textbf{3.88} & 2.73 / 5.61 \\
    Close Drawer & \textbf{5.80} / --- & \underline{5.80} / --- & 6.11 / --- \\
    \bottomrule
  \end{tabular}
\end{table}

\begin{table}[t]
  \centering
  \caption{\textbf{Task-wise whole-body motion quality for motion
  initialization.} Metrics are averaged over the same fixed random 25-trajectory
  cohort per task as
  Table~\ref{tab:initialization}, using the definitions in
  Appendix~\ref{sec:motion-quality-metrics}. Lower is better. Bold and underline
  mark the best and second-best values per task.}
  \label{tab:motion-quality-taskwise}
  \papertablestyle
  \begin{tabular}{@{}llcccc@{}}
    \toprule
    Task & Method & Pelvis drop (cm) & Tilt ($^\circ$) & Joint step ($^\circ$)
      & Extra yaw ($^\circ$) \\
    \midrule
    \multirow{3}{*}{Close Laptop}
      & SONIC Rollout & \textbf{0.28} & \textbf{9.62} & \textbf{2.87} & \underline{3.94} \\
      & Retargeting & \underline{0.50} & \underline{10.86} & \underline{3.95} & \textbf{3.67} \\
      & Whole-Body IK & 2.06 & 11.32 & 6.07 & 22.60 \\
    \cmidrule(lr){1-6}
    \multirow{3}{*}{Open Fridge}
      & SONIC Rollout & \underline{3.51} & \textbf{10.39} & \textbf{7.79} & \textbf{7.39} \\
      & Retargeting & \textbf{3.35} & \underline{11.62} & \underline{8.28} & \underline{13.96} \\
      & Whole-Body IK & 8.08 & 13.04 & 8.42 & 16.70 \\
    \cmidrule(lr){1-6}
    \multirow{3}{*}{Pull Curtain}
      & SONIC Rollout & \textbf{1.94} & \textbf{5.32} & \underline{5.31} & \underline{6.55} \\
      & Retargeting & \underline{2.48} & \underline{6.16} & \textbf{5.26} & \textbf{5.66} \\
      & Whole-Body IK & 2.92 & 7.14 & 5.92 & 7.24 \\
    \cmidrule(lr){1-6}
    \multirow{3}{*}{Close Drawer}
      & SONIC Rollout & \underline{7.40} & \underline{19.22} & \textbf{7.78} & \textbf{10.84} \\
      & Retargeting & \textbf{6.87} & \textbf{18.83} & \underline{8.58} & \underline{22.51} \\
      & Whole-Body IK & 7.95 & 19.57 & 10.60 & 48.16 \\
    \bottomrule
  \end{tabular}
\end{table}

\subsection{Refinement Depth and Processing Cost}
\label{sec:refinement-depth-details}
This study measures the accuracy obtained at a specified optimization depth
and the processing time needed to reach it. In
Figure~\ref{fig:kd-iterations}, $K^mD^k$ denotes $m$ kinematic correction
rounds followed by $k$ dynamics-aware refinement rounds, starting from the
SONIC Rollout reference.

\paragraph{Depth comparisons.}
We compare $D^8$ with $KD^8$ and extend the $D$-only branch to 16 rounds.
Repeated-$K$ comparisons use $m\in\{1,2,4,8\}$, both alone and followed by
eight $D$ rounds. The $KD^k$ curve reports the raw output at each depth
$k=0,\ldots,8$. Thus, these curves measure the effect of additional
optimization without historical candidate selection.

\paragraph{Time accounting.}
Table~\ref{tab:alignment-cost} pairs processing times with the endpoint
errors in Figure~\ref{fig:kd-iterations}. Elapsed time runs from job start
to capture completion, including preparation, optimization, and replay.
When an experiment reuses an existing initialization ($H$) or kinematic
correction ($K$), its recorded processing time is included in the total. Queueing,
discarded failed attempts, and subsequent candidate selection and validation
are excluded. Episode times are normalized by frame count, averaged within
tasks, and then averaged equally across tasks.

\paragraph{Hardware and interpretation.}
The logged runs use workstations with Ryzen 9 9950X3D CPUs and RTX 4090 GPUs
(48\,GB VRAM), and a machine with a Core Ultra 9 275HX CPU and an RTX 5090
Laptop GPU (24\,GB VRAM). Each host is configured to process three episodes concurrently,
while $K/D$ rounds within an episode remain sequential. Times are pooled
across hosts without division by the number of concurrent episodes, so they characterize these
processing runs rather than single-machine latency. Under this accounting,
$K^1D^8$ costs 1.49\,s/frame; increasing $K$ depth to eight raises the cost
to 3.38\,s/frame with similar endpoint error.

\begin{table}[t]
  \centering
  \caption{\textbf{Processing cost and endpoint accuracy.}
  Representative configurations from Figure~\ref{fig:kd-iterations}.
  Time includes preparation, optimization, and replay, with reused $H/K$
  costs included. It excludes queueing, discarded failed attempts, and
  subsequent selection and validation. Values are equal-task means;
  orientation averages the three hand tasks. Errors use raw outputs
  at each depth (Figure~\ref{fig:kd-iterations}).}
  \label{tab:alignment-cost}
  \papertablestyle
  \setlength{\tabcolsep}{8pt}
  \begin{tabular}{lccc}
    \toprule
    Configuration & Time (s/frame) & Pos. (cm) & Ori. ($^\circ$) \\
    \midrule
    $K^1$ & 0.65 & 4.86 & 16.46 \\
    $K^8$ & 2.55 & 4.77 & 16.51 \\
    $D^8$ & 1.27 & 8.40 & 16.40 \\
    $D^{16}$ & 2.11 & 7.16 & 12.36 \\
    $K^2D^8$ & 1.80 & 4.04 & 7.90 \\
    $K^4D^8$ & 2.33 & 4.11 & 7.94 \\
    $K^8D^8$ & 3.38 & 4.01 & 7.90 \\
    \midrule
    \rowcolor[HTML]{EDF5F3}
    $K^1D^8$ (Ours) & 1.49 & 3.97 & 8.08 \\
    \bottomrule
  \end{tabular}
\end{table}

\subsection{Response-Model Ablation}
\label{sec:response-ablation}
This ablation isolates the learned execution response from the rest of the
refinement procedure. Table~\ref{tab:response-model} compares the learned
response (Appendix~\ref{sec:response-training}) with a unit response, which
assumes that the executed endpoint change equals the change predicted by
the kinematic Jacobian. Both use the same Jacobian and token optimizer.
Both conditions share $H/K$ inputs, targets, refinement settings, and eight
$D$ rounds; endpoint configurations remain fixed across episodes.

Position and orientation errors are taken from the same historically
selected candidate.

\subsection{Action Alignment across 50 Tasks}
\label{sec:expanded-task-benchmark}
\paragraph{Scope and protocol.}
The expanded simulation study tests alignment across 50 tasks using
112 paired trajectories, separate from the four-task cohort. Stage,
refinement-depth, and response comparisons use this same cohort and equal
task weights. Stage comparisons select from all recorded candidates; depth
and response comparisons use the output of each specified round.

\paragraph{Kinematic correction and dynamics-aware refinement.}
Table~\ref{tab:expanded-stage-comparison} extends the stage-wise gains in
Table~\ref{tab:action-transfer}: $K$ reduces position/orientation errors from
17.03~cm/36.06$^\circ$ to 5.22~cm/21.81$^\circ$; refinement and historical
selection further reduce them to 3.83~cm/10.85$^\circ$.

\begin{table}[!htbp]
  \centering
  \caption{\textbf{Stage comparison on the expanded benchmark.}
  Task-equal endpoint errors over 112 paired trajectories across 50 tasks;
  lower is better.
  $H+K+D_s$ uses historical candidate selection.}
  \label{tab:expanded-stage-comparison}
  \papertablestyle
  \begin{tabular}{lcc}
    \toprule
    Stage & Pos. (cm) & Ori. ($^\circ$) \\
    \midrule
    $H$ & 17.03 & 36.06 \\
    $H+K$ & 5.22 & 21.81 \\
    $H+K+D_s$ & \textbf{3.83} & \textbf{10.85} \\
    \bottomrule
  \end{tabular}
\end{table}

\begin{figure}[!htbp]
  \centering
  \includegraphics[width=\linewidth]{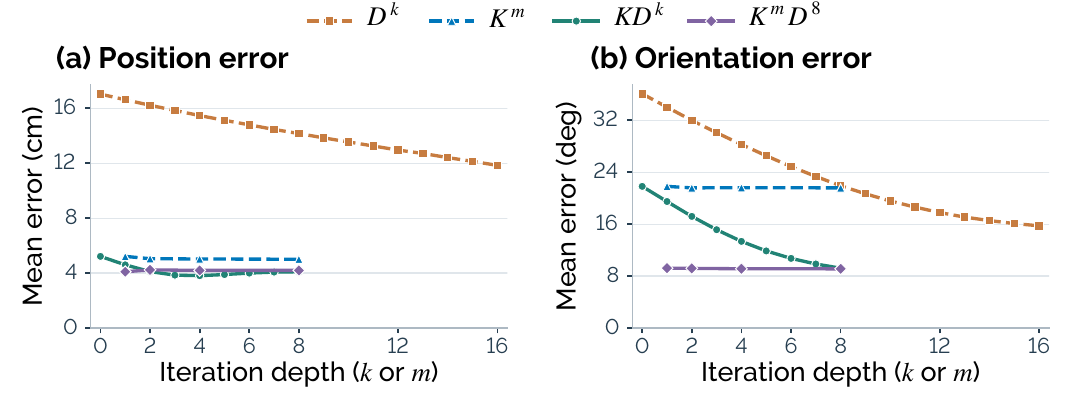}
  \caption{\textbf{Refinement depth on the expanded benchmark.}
  Task-equal averages of raw stage outputs for $D^k$, repeated $K^m$,
  $KD^k$, and $K^mD^8$ on paired trajectories.}
  \label{fig:benchmark-refinement-depth}
\end{figure}

\paragraph{Refinement depth.}
Figure~\ref{fig:benchmark-refinement-depth} compares candidates after the
same number of refinement rounds, without selecting earlier candidates.
Increasing $K$ depth from one to eight
changes position error from 5.22 to 5.01~cm. One $K$ round followed by
eight $D$ rounds reaches 4.10~cm and 9.21$^\circ$, compared with
5.22~cm and 21.81$^\circ$ after $K$ alone. Eight $D$ rounds without
$K$ yield 14.16~cm and 21.91$^\circ$.
Repeated $K$ offers little gain, while combining $K$ and $D$ lowers errors further.

\paragraph{Learned execution response.}
After one $K$ and eight $D$ rounds, the MLP lowers position error from
4.25 to 4.10~cm and orientation error from 9.39$^\circ$ to 9.21$^\circ$
(Table~\ref{tab:expanded-response-model}).

\begin{table}[!htbp]
  \centering
  \caption{\textbf{Response-model comparison on the expanded benchmark.}
  Task-equal errors over 112 trajectories across 50 tasks after $KD^8$,
  without historical selection. Lower is better.}
  \label{tab:expanded-response-model}
  \papertablestyle
  \begin{tabular}{lcc}
    \toprule
    Response & Pos. (cm) & Ori. ($^\circ$) \\
    \midrule
    Unit response & 4.25 & 9.39 \\
    MLP response & \textbf{4.10} & \textbf{9.21} \\
    \bottomrule
  \end{tabular}
\end{table}

\section{Online IK under a Whole-Body Tracking Controller}
\label{sec:online-ik-instability}
This study examines whether online IK can correct endpoints while preserving
coordinated whole-body execution under SONIC. We compare three ways of
combining IK and SONIC joint commands on five trajectories per task, giving 60
runs across four tasks.
This cohort is separate from Appendix~\ref{sec:action-ablation-protocol}.
The IK objective includes position and orientation, with orientation weight
0.01. Runs are neither accuracy-filtered nor retried for better outcomes.

\subsection{Three Command Compositions}
\label{sec:online-ik-variants}
The variants change which joints receive IK commands and whether SONIC's
balance contribution is added. Table~\ref{tab:online-ik-compositions}
defines the same three choices for hand and foot interaction. Variant A
uses the broader IK joint set without the balance contribution; B adds
that contribution to A; C restricts IK to fewer joints and leaves the
remaining joints under SONIC control. Balance and contact forces dynamically couple the base and limbs,
including those with unchanged commands.

\begin{table}[!htbp]
\centering
\caption{Joint-command assignments for the three online-IK variants.}
\label{tab:online-ik-compositions}
\papertablestyle
\begin{tabular}{@{}lp{0.37\linewidth}p{0.37\linewidth}@{}}
\toprule
Variant & Three hand tasks & Foot-operated Close Drawer \\
\midrule
A & IK controls arms and waist; SONIC controls legs.
& IK controls both legs; SONIC controls remaining joints. \\
B & Same as A, with SONIC's balance contribution added.
& Same as A, with SONIC's balance contribution added. \\
C & IK controls arms; SONIC controls waist and legs.
& IK controls the right operating leg; SONIC controls remaining joints. \\
\bottomrule
\end{tabular}
\end{table}

\subsection{Accuracy and Stability}
\label{sec:online-ik-results}
We evaluate endpoint accuracy and stability together because reaching the
final replay frame does not establish stable execution.
Table~\ref{tab:online-ik-stability} averages tracking errors over frames,
both endpoints, and episodes; tilt follows the episode-maximum definition
in Appendix~\ref{sec:motion-quality-metrics}. Runs that fail during initialization are counted separately because they
do not reach the trajectory replay used to measure tracking error. Fall counts include repeated events within a run.

\begin{table}[!htbp]
  \centering
  \caption{\textbf{Online-IK stability diagnostics on five episodes per task.}
  Errors use measured endpoints, not IK solutions.
  Init. falls counts failed runs; Events counts logged falls, including
  startup. Dashes indicate no formal replay.
  Bold/underline rank values within each task from lowest to second lowest;
  they do not indicate that execution is stable.}
  \label{tab:online-ik-stability}
  \papertablestyle
  \begin{tabular}{@{}llrrrrr@{}}
    \toprule
    Variant & Task & Pos. (cm) & Ori. ($^\circ$) & Tilt ($^\circ$) & Init. falls & Events \\
    \midrule
    A & Close Laptop & \underline{4.45} & \underline{10.43} & \underline{16.97} & 0/5 & \textbf{0} \\
     & Open Fridge & \underline{2.33} & \underline{15.34} & \underline{18.97} & 0/5 & \textbf{0} \\
     & Pull Curtain & \underline{1.58} & \underline{10.17} & \underline{7.03} & 0/5 & \textbf{0} \\
     & Close Drawer & --- & --- & --- & 5/5 & --- \\
    \midrule
    B & Close Laptop & \textbf{2.43} & \textbf{8.28} & \textbf{13.30} & 0/5 & \textbf{0} \\
     & Open Fridge & \textbf{1.55} & \textbf{14.45} & \textbf{17.26} & 0/5 & \textbf{0} \\
     & Pull Curtain & \textbf{1.50} & \textbf{10.07} & \textbf{6.61} & 0/5 & \textbf{0} \\
     & Close Drawer & --- & --- & --- & 5/5 & --- \\
    \midrule
    C & Close Laptop & 26.50 & 40.93 & 82.75 & 0/5 & \underline{9} \\
     & Open Fridge & 10.61 & 20.55 & 34.35 & 0/5 & \underline{1} \\
     & Pull Curtain & 9.00 & 18.66 & 10.87 & 0/5 & \textbf{0} \\
     & Close Drawer & --- & --- & --- & 5/5 & --- \\
    \bottomrule
  \end{tabular}
\end{table}

\paragraph{Observed behavior and scope.}
Video inspection identifies instability in all 60 runs. All 15 drawer runs
fall during first-pose preparation, whereas the 45 hand-task runs reach
their final frames without stable tracking. Variant C logs nine laptop
fall events and one fridge fall event. Variant B achieves hand-task
position errors of 1.50--2.43\,cm but remains visibly unstable, showing why
endpoint error alone is insufficient for accepting these trajectories.
We therefore exclude the tested variants from policy training. This finding
is limited to these SONIC implementations.

\subsection{Qualitative Failure Sequences}
\label{sec:online-ik-sequences}
Figures~\ref{fig:online-ik-falls}--\ref{fig:online-ik-falls-C} complement
the aggregate diagnostics with one simulation sequence per implementation
and task. Each sequence is chosen by peak base tilt across the five runs,
including initialization. These sequences illustrate failure patterns, not their frequency.

\newcommand{\onlineikfigure}[3]{%
\begin{figure}[p]
  \centering
  \begingroup
  \setlength{\tabcolsep}{2pt}
  \newcommand{\iksequence}[1]{%
    \includegraphics[width=.24\linewidth]{figures/online_ik/variant_#1_##1_1.jpg}&%
    \includegraphics[width=.24\linewidth]{figures/online_ik/variant_#1_##1_2.jpg}&%
    \includegraphics[width=.24\linewidth]{figures/online_ik/variant_#1_##1_3.jpg}&%
    \includegraphics[width=.24\linewidth]{figures/online_ik/variant_#1_##1_4.jpg}}
  \begin{tabular}{@{}cccc@{}}
    \multicolumn{4}{@{}l}{{\fontfamily{Raleway-TLF}\selectfont\bfseries (a) Close Laptop}} \\
    \iksequence{close_laptop} \\[3pt]
    \multicolumn{4}{@{}l}{{\fontfamily{Raleway-TLF}\selectfont\bfseries (b) Open Fridge}} \\
    \iksequence{open_fridge} \\[3pt]
    \multicolumn{4}{@{}l}{{\fontfamily{Raleway-TLF}\selectfont\bfseries (c) Pull Curtain}} \\
    \iksequence{close_curtain} \\[3pt]
    \multicolumn{4}{@{}l}{{\fontfamily{Raleway-TLF}\selectfont\bfseries (d) Close Drawer}} \\
    \iksequence{close_drawer}
  \end{tabular}
  \endgroup
  \caption{\textbf{Online-IK instability: variant #1 (#2).}
  Each row follows one continuous segment from left to right, selected by
  peak base tilt as described in Appendix~\ref{sec:online-ik-sequences}.
  A fixed crop within each sequence preserves relative motion and keeps
  robot scale comparable across tasks. Close Drawer ends in an
  initialization fall. Task objects are omitted.}
  \label{#3}
\end{figure}
}

\onlineikfigure{A}{IK without balance overlay}{fig:online-ik-falls}

\onlineikfigure{B}{IK with SONIC balance overlay}{fig:online-ik-falls-B}

\onlineikfigure{C}{restricted IK joint control}{fig:online-ik-falls-C}

\end{document}